\documentclass[10pt,journal,cspaper]{IEEEtran}

\usepackage[ruled]{algorithm2e}
\usepackage[font=small]{caption}
\usepackage{subfigure}
\usepackage{booktabs}
\usepackage{bigstrut}
\usepackage{enumerate}
\usepackage{cite}
\usepackage{graphicx}
 \usepackage{epstopdf}
 \graphicspath{{../pdf/}{../jpeg/}}
\usepackage{amsmath,amssymb,amsthm,mathrsfs,amsfonts,dsfont}
\usepackage{bm}
\usepackage{algorithmic}
\usepackage{mdwmath}
\usepackage{mdwtab}
\usepackage{array}
\usepackage{tikz}
\usetikzlibrary{positioning}
\usetikzlibrary{shapes,snakes}

\makeatletter

\newcommand{\Rmnum}[1]{\expandafter\@slowromancap\romannumeral #1@}
\makeatother

\newtheorem{plain}{Property}
\newtheorem{definition}{Definition}

\begin{document}

\title{Nonparametric Relational Topic Models through Dependent Gamma Processes}

\author{Junyu~Xuan,~Jie Lu,~\IEEEmembership{Senior Member,~IEEE},~Guangquan Zhang,\\~Richard~Yi~Da~Xu,~and~Xiangfeng~Luo,~\IEEEmembership{Member,~IEEE}
\IEEEcompsocitemizethanks{
\IEEEcompsocthanksitem J. Xuan is with the Centre for Quantum Computation and Intelligent Systems (QCIS), School of Software, Faculty of Engineering
and Information Technology, University of Technology Sydney (UTS), Australia and the School of Computer Engineering and Science, Shanghai University, China (e-mail: Junyu.Xuan@student.uts.edu.au).
\IEEEcompsocthanksitem J. Lu is with Faculty of Engineering
and Information Technology, University of Technology Sydney (UTS),
Australia (e-mail: Jie.Lu@uts.edu.au).
\IEEEcompsocthanksitem G. Zhang is with the Centre for Quantum Computation and
Intelligent Systems (QCIS), School of Software, Faculty of Engineering
and Information Technology, University of Technology Sydney (UTS),
Australia (e-mail: Guangquan.Zhang@uts.edu.au).
\IEEEcompsocthanksitem R. Y. D. Xu is with Faculty of Engineering
and Information Technology, University of Technology Sydney (UTS),
Australia (e-mail: Yida.Xu@uts.edu.au).
\IEEEcompsocthanksitem X. Luo is with the School of Computer Engineering and Science, Shanghai University, China. (e-mail: luoxf@shu.edu.cn).
}
\thanks{}}


\IEEEcompsoctitleabstractindextext{%
\begin{abstract}
Traditional Relational Topic Models provide a way to discover the hidden topics from a document network. Many theoretical and practical tasks, such as dimensional reduction, document clustering, link prediction, benefit from this revealed knowledge.
However, existing relational topic models are based on an assumption that the number of hidden topics is known in advance, and this is impractical in many real-world applications. Therefore, in order to relax this assumption, we propose a nonparametric relational topic model in this paper. Instead of using fixed-dimensional probability distributions in its generative model, we use stochastic processes. Specifically, a gamma process is assigned to each document, which represents the topic interest of this document. Although this method provides an elegant solution, it brings additional challenges when mathematically modeling the inherent network structure of typical document network, i.e., two spatially closer documents tend to have more similar topics. Furthermore, we require that the topics are shared by all the documents. In order to resolve these challenges, we use a subsampling strategy to assign each document a different gamma process from the global gamma process, and the subsampling probabilities of documents are assigned with a Markov Random Field constraint that inherits the document network structure. Through the designed posterior inference algorithm, we can discover the hidden topics and its number simultaneously. Experimental results on both synthetic and real-world network datasets demonstrate the capabilities of learning the hidden topics and, more importantly, the number of topics.
\end{abstract}

\begin{keywords}
Topic models, Nonparametric Bayesian learning, Gamma process, Markov random field
\end{keywords}}

\maketitle

\IEEEdisplaynotcompsoctitleabstractindextext


\section{Introduction}

\IEEEPARstart{U}{nderstanding} a corpus is significant for businesses, organizations and individuals for instance the academic papers of IEEE, the emails in an organization and the previously browsed webpages of a person. One commonly accepted and successful way to understand a corpus is to discover the hidden topics in the corpus \cite{blei2003latent,Blei:2012:PTM}. The revealed hidden topics could improve the services of IEEE, such as the ability to search, browse or visualize academic papers ; help an organization understand and resolve the concerns of its employees; help internet browsers understand the interests of a person and then provide accurate personalized services. Furthermore, there are normally links between the documents in a corpus.
A paper citation network \cite{6494572} is an example of a document network in which the academic papers are linked by their citation relations; an email network \cite{klimt2004enron} is a document network in which the emails are linked by their reply relations; a webpage network \cite{park2003hyperlink} is a document network in which webpages are linked by their hyperlinks. Since these links also express the nature of the documents, it is apparent that hidden topic discovery should consider these links as well.

Similar studies focusing on the hidden topics discovering from the document network using some Relational Topic Models (RTM) \cite{chang2009relational,chang2010hierarchical,zhujun2013} have already been successfully developed. Unlike the traditional topic models \cite{blei2003latent,Blei:2012:PTM} that focus on mining the hidden topics from a document corpus (without links between documents), the RTM can make discovered topics inherit the document network structure. The links between documents can be considered as constrains of the hidden topics.

One drawback of existing RTMs is that they are built with fixed-dimensional probability distributions, such as Dirichlet, Multinomial, Gamma and Possion distribution, which require their dimensions be fixed before use. Hence, the number of hidden topics must be specified in advance, and is normally chosen using domain knowledge. This is difficult and unrealistic in many real-world applications, so RTMs fail to find the number of topics in a document network.

In order to overcome this drawback, we propose a Nonparametric Relational Topic (NRT) model in this paper, which removes the necessity of fixing the topic number. Instead of probability distributions, stochastic processes are adopted by the proposed model. Stochastic process can be simply considered as `infinite' dimensional distributions\footnote{We only consider the pure-jump processes in this paper. Some continuous processes cannot be simply considered as the `infinite' dimensional distributions.}. In order to express the interest of a document on the `infinite' number of topics, we assign each document a Gamma process that has infinite components. An additional requirement for the Gamma process assignment is that the two linked documents should have a tendency to share similar topics. This is a common feature found in many real-world applications, and many literatures \cite{chang2009relational,chang2010hierarchical,zhujun2013} have exploited this property in their work.
In order to achieve the above requirement, we have formally defined two properties that any relational topic model of a document network should satisfy. First we use a global gamma process to represent a set of base components that is shared by all documents. This is important because users are not interested in analyzing documents in a database without sharing any common topics \cite{teh2006hierarchical}. Our model achieves the defined properties through: 1) thinning the global gamma process with document-dependent probabilities; 2) adding a Markov Random Field constraint to the thinning probabilities to retain the network structure. Finally, we assign each document with a gamma process that inherits both the content of the document and the link structure. Two sampling algorithms are designed to learn the proposed model under different conditions. Experiments with document networks show some efficiency in learning what the hidden topics are and superior performance the model's ability to learn the number of hidden topics. It is worth noting that, although we use document networks as examples throughout this paper, our work can be applied to other networks with node features.

The main contributions of this paper are to:
\begin{enumerate}
  \item propose a new nonparametric Bayesian model which can relax the topic number assumption used in the traditional relational topic models;
  \item design two sampling inference algorithms for the proposed model: a truncated version and an exact version.
\end{enumerate}

The rest paper is structured as follows. Section \Rmnum{2} summarizes the related work. The proposed NRT model is presented in Section \Rmnum{3} and we have illustrated the detailed derivations of its sampling inference in Section \Rmnum{4}. Section \Rmnum{5} presents experimental results both on the synthetic and real-world data. Finally, Section \Rmnum{6} concludes this study with a discussion on future directions.


\section{Related Work}

In this section, we briefly review the related work of this paper. The first part summarizes the literature on relational topic models. The second part summarizes the literatures on nonparametric Bayesian learning.

\subsection{Topic Models with network}

%

Our work in this paper aims to model the data with the network structure as a constraint. Since social network and citation network are two explicit and commonly-used networks in the data mining and machine learning areas, some extensions of the traditional topic models try to adapt to these networks. For the social network, an Author-Recipient-Topic model \cite{mccallum2007topic} was proposed to analyze the categories of roles in social networks based on the relationships of people in the network. A similar task was investigated in \cite{cha2012social} where social network structure was inferred from informal chat-room conversations utilizing the topic model \cite{tuulos2004combining}. The `noisy links' and `popularity bias' of social network was addressed by a properly designed topic model in \cite{wang2011dynamic} and \cite{cha2013incorporating}. As an important issue of social network analysis, communities \cite{mei2008topic} were extracted using a Social Topic Model \cite{pathak2008social}. The Mixed Membership Stochastic Blockmodel is another way to learn the mixed membership vector (i.e., topic distribution) for each node from a network structure \cite{airoldi2009mixed}, but it did not consider the content/features of each node. For the citation network, Relational Topic Model (RTM) was proposed to infer the topics \cite{chang2009relational}, discriminative topics \cite{zhujun2013} and hierarchical topics \cite{chang2010hierarchical} from citation networks by introducing a link variable between two linked documents. Unlike RTM, a block was adopted to model the link between two document \cite{nallapati2008joint,zhu2013scalable}. Considering the physical meaning of citation relations, a variable was introduced to indicate if the content of citing paper was inherited from cited paper or not \cite{dietz2007unsupervised,he2009detecting}. In order to keep the document structure, Markov Random Field (MRF) was combined with topic model \cite{sun2009itopicmodel}. The communities in citation network were also investigated \cite{liu2009topic}.

In summary, existing relational topic models are all inherited from traditional topic models, so the number of topics needs to be fixed. It is unrealistic, in many real-world situations, to fix this number in advance. Our work tries to resolve this issue through the nonparametric learning techniques reviewed in the following subsection.

\subsection{Relational Topic Model}

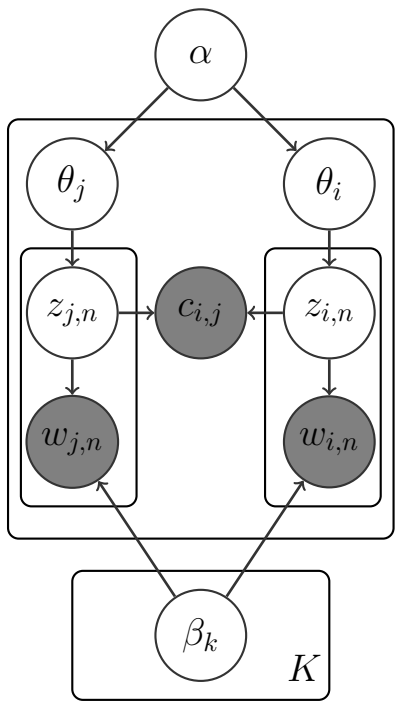
\begin{figure}[t]
\centering
    \tikzstyle{rv}=[circle,
                        thick,
                        minimum size=1.2cm,
                        draw=black!80,
                        ]
    \tikzstyle{line}=[->,
                        solid,
                        line width=1pt,
                        draw=black!80,
                        ]
 \tikzstyle{dt}=[circle,
                        thick,
                        minimum size=1.2cm,
                        draw=black!80,
                        fill=gray
                        ]
    \begin{tikzpicture}[font=\Large,scale=0.85]
     	

    	\node[rv] (1) at (0, 0) {$\alpha$};

    	\draw[thick, rounded corners] (-3,-7.5) rectangle (3,-1);

        \node[dt] (2) at (0, -4) {$c_{i,j}$};

    	\node[rv] (3) at (2, -2) {$\theta_{i}$};
        \draw[thick, rounded corners] (-2.8,-7) rectangle (-1,-3);
    	\node[rv] (4) at (2, -4) {$z_{i, n}$};
        \node[dt] (5) at (2, -6) {$w_{i, n}$};

        \node[rv] (6) at (-2, -2) {$\theta_{j}$};
        \draw[thick, rounded corners] (2.8,-7) rectangle (1,-3);
    	\node[rv] (7) at (-2, -4) {$z_{j, n}$};
        \node[dt] (8) at (-2, -6) {$w_{j, n}$};

        \draw[thick, rounded corners] (-2,-10) rectangle (2,-8);
        \node[rv] (9) at (0, -9) {$\beta_k$};

        \node at (1.2,-9.5) [right] {$K$};

    	\path[line]
    		(1)		edge (3)
    		(1)		edge (6)
            (3)		edge (4)
    		(4)		edge (5)
            (6)		edge (7)
    		(7)		edge (8)
            (4)		edge (2)
    		(7)		edge (2)
            (9)		edge (5)
    		(9)		edge (8)
            ;

    \end{tikzpicture}
\caption{Finite Relational-Topic-Model}
  \label{frtm}
\end{figure}

Since the finite relational topic models are our comparative model, we introduce a relational topic model \cite{chang2009relational} here in detail. The corresponding graphical representation is shown in Fig. \ref{frtm}, and the generative process is as follows,
\begin{equation}
\begin{aligned}
\theta_d &\overset{i.i.d}{\sim} Dirichlet(\alpha)\\
\phi_k &\overset{i.i.d}{\sim} Dirichlet(\beta)\\
z_{d, n} &\sim Discrete(\theta_d)\\
w_{d, n} &\sim Discrete(\phi_{z_{d,n}})\\
c_{d_i, d_j} &\sim GLM(z_{d_i}, z_{d_j})
\end{aligned}
\label{rtm1}
\end{equation}
where $\theta_d$ is the topic distribution of a document, $\phi_k$ is the word distribution of topic $k$, $z_{d,n}$ is the topic index of word $n$ in document $d$, $w_{d,n}$ is observed word $n$ in document $d$. All these variables are same with the original LDA \cite{blei2003latent}. The different and significant part is the variable $c$, which denotes the observed document link. This model uses a Generalized Linear Model \cite{mccullagh1984generalized} to model the generation of the document links.
\begin{equation}
\begin{aligned}
p(c_{d_i, d_j} = 1) &\sim GLM(z_{d_i}, z_{d_j})
\end{aligned}
\end{equation}
One problem with this model is that the number of topics needs to be pre-defined and for some real-world applications, this is not trivial.

\subsection{Nonparametric Bayesian Learning}

Nonparametric Bayesian learning \cite{gershman2012tutorial} is a key approach for learning the number of mixtures in a mixture model (also called the model selection problem). Without predefining the number of mixtures, this number is supposed to be inferred from the data, i.e., let the data speak.

The traditional elements of probabilistic models are fixed-dimensional distributions, such as Gaussian distribution, Dirichlet distribution \cite{blei2003latent}, Logistic Normal distribution \cite{blei2007correlated}, and so on. All these distributions need to predefine their dimensions. In order to avoid this, Gaussian process \cite{seeger2004gaussian} and Dirichlet process \cite{ghosal2010dirichlet} are used to replace former fixed-dimensional distributions because of their infinite properties. Since the data is limited, the learned/used atoms will also be limited even with these `infinite' stochastic processes.

Dirichlet Process can be seen as a distribution over distributions. Since a sample from Dirichlet Process defines a bunch of variables that satisfies Dirichlet distribution, Dirichlet process is a good alternative for the models with Dirichlet distribution as the prior. There are three different methods to construct this process: Black-MacQueen urn schema \cite{blackwell1973ferguson}, Chinese restaurant process \cite{teh2010dirichlet} and stick breaking process \cite{teh2010dirichlet}. Although the processes that result from them are all Dirichlet processes, they can express different properties of Dirichlet process, such as the posterior distribution from Black-MacQueen urn schema, the clustering from Chinese restaurant process and the formal sampling function from stick breaking process. Based on these constructive processes, a Dirichlet process mixture \cite{antoniak1974mixtures} is proposed, which is a kind of infinite mixture models. Infinite mixture models are the extension of Finite Mixture Models where there are a finite number of hidden components (topics) used to generate data. Another infinite mixture model is the Infinite Gaussian mixture model \cite{rasmussen1999infinite}. Normally, a Gaussian mixture model is used for continuous variables and a Dirichlet process mixture is used for discrete variables. An example use for a Dirichlet process is the hierarchical topic model composed by Latent Dirichlet Allocation (LDA) \cite{blei2003latent} with a nested Chinese restaurant process \cite{griffiths2004hierarchical}. By using a nested Chinese restaurant process as the prior, not only is the number of them not fixed, the topics in this model are also hierarchically organized. In order to learn the Dirichlet process mixture based models with an infinite property, the inference methods should be properly designed. There are two popular and successful methods to do this: Markov Chain Monte Carlo (MCMC) \cite{neal2000markov} and variational inference \cite{carin2011variational}.


To summarize, nonparametric learning has been successfully used for extending many models and applied in many real-world applications. However, there is still no work on the nonparametric extension of relational topic models. This paper uses a set of Gamma processes to extend the finite relational topic model to the infinite one.


%
%
%
%
%
%


\section{Nonparametric Relational Topic Model}

In this section, we present the proposed Nonparametric Relational Topic (NRT) model in detail. This model can handle the issue that the number of topics needs to be defined.

The proposed model uses a Gamma process to express the interest of a document on infinite hidden topics. A gamma process $GaP(\alpha, H)$ \cite{gp2014} is a stochastic process, where $H$ is a base (shape) measure parameter and $\alpha$ is the concentration (scale) parameter. It also corresponds to a complete random measure. Let $\Gamma = \{ (\pi_i, \theta_i)\}_{i=1}^{\infty}$ be a random realization of a Gamma process in the product space $\mathds{R}^+ \times \Theta$. Then, we have
\begin{equation}\label{gammap}
\begin{aligned}
\Gamma &\sim GaP(\alpha, H) \\
&= \sum_{i=1}^{\infty} \pi_i \delta_{\theta_i}
\end{aligned}
\end{equation}
where $\delta(\cdot)$ is an indicator function, $\pi_i$ satisfies an improper gamma distribution with parameters $gamma(0, \alpha)$ and $\theta_i \sim H$.
When using $\Gamma$ to express the document interest, the $\{\theta_i)\}_{i=1}^{\infty}$ in Eq. (\ref{gammap}) denotes the infinite number of topics and $\{\pi_i)\}_{i=1}^{\infty}$ in Eq. (\ref{gammap}) denotes the weights of infinite number of topics in a document.

\begin{figure}[t]
  \centering
  \includegraphics[scale=0.3]{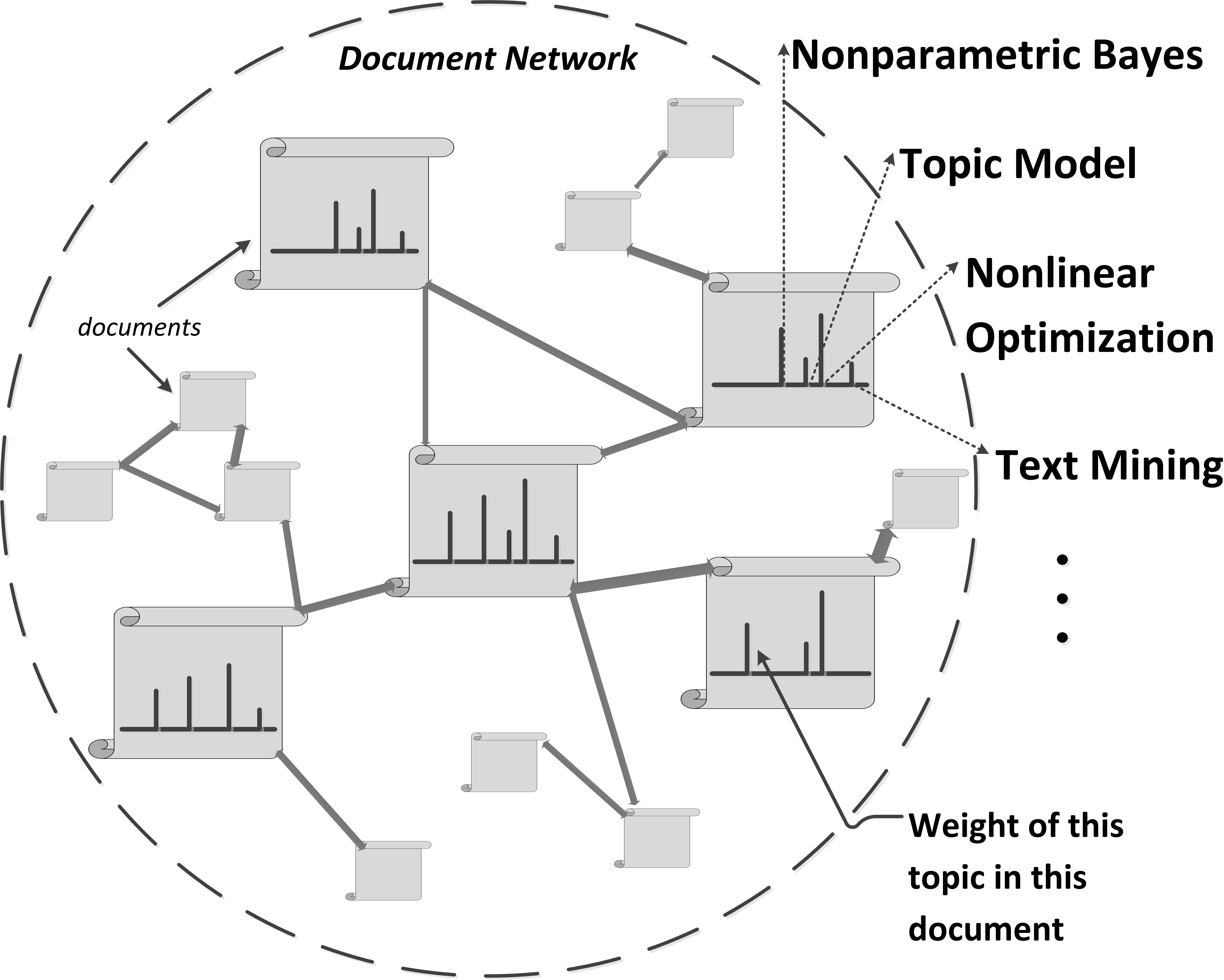}
  \caption{Illustration of Gamma process assignments for the document network. Each document is assigned a Gamma process which has infinite components (represented by the fences in a document in the figure). Each fence denotes a hidden topic, and some examples are shown in the figure. The length of the fences denote the weights of different topics in a document.}
  \label{illupic}
\end{figure}

As illustrated in Fig. \ref{illupic}, our idea is to assign each document a gamma process. This assignment should satisfy the following two properties:

\begin{plain}\label{p1} Two Gamma processes of two documents with a link should have similar components/topics with each other.
\end{plain}

\begin{plain}\label{p2} All the Gamma processes of documents should share the same set of components/topics.
\end{plain}

In order to achieve the above properties, we firstly generate a global Gamma process,
\begin{equation*}
\begin{aligned}
\Gamma_0 &\sim GaP(\alpha, H)
\end{aligned}
\end{equation*}
which is equal to
\begin{equation*}
\begin{aligned}
\Gamma_0 &= \sum_{k=1}^{\infty} \pi_k \delta_{(x_k, \theta_k)},~~~\theta_k \sim H
\end{aligned}
\end{equation*}
where $\{\pi_k, \theta_k\}_{k=1}^\infty$ is the shared global set of components/topics for documents. We then hope the components of the Gamma process for each document falls within the set of components/topics in the global Gamma process. We use a dependent thinned gamma process to achieve this goal. Its definition is as follow,
\begin{definition}[Thinned Gamma Process \cite{foti2012unifying}] Suppose we have a gamma process $\Gamma \sim GaP(\alpha, H)$ and we know there are countably infinite points $\{ (\pi_i, \theta_i)\}_{i=1}^{\infty}$. Then, we generate a set of independent binary variables $\{r\}_{i=1}^{\infty}$ ($r_i \in \{0, 1\}$). The new process,
\begin{equation}\label{1}
\begin{aligned}
\Gamma_{1} = \sum_{i=1}^{\infty} \pi_i r_i \delta_{\theta_i}
\end{aligned}
\end{equation}
is still a gamma process, which is proofed by \cite{foti2012unifying}. The $\{r_i\}$ can be seen as the indicators for the reservation of the point of original/global gamma process, so $\Gamma_{1}$ is called \emph{thinned gamma process}.
\end{definition}

We can give each $r_i$ a Bernoulli prior $p(r_i = 1) = p_i$. Apparently, different realizations of $\{r_i\}$ will lead to different gamma processes. Furthermore, the dependence between the different realizations of $\{r_i\}$ will also lead to dependence of the generated gamma processes.

For each document, a thinned gamma process $\Gamma_d$ is generated with $\Gamma_0$ as the global process,
\begin{equation}
\begin{aligned}
\Gamma_{d} = \sum_{k=1}^{\infty} \pi_k r^d_k \delta_{\theta_k}
\end{aligned}
\end{equation}
where $\{r^d_k\}_{k=1}^{\infty}$ is a set of indicators of document $d$ on the corresponding components. These $\{r^d_k\}_{k=1}^{\infty}$ are independent identical distributed random variables with Bernoulli distributions,
\begin{equation}
\begin{aligned}
r^d_k \sim Bernoulli(q_k^d)
\end{aligned}
\end{equation}
where $q_k^d$ denotes the probability of the Gamma process $\Gamma_d$ of document $d$ with component $k$. Therefore, Property \ref{p2} is achieved.

In order to make the linked documents have similar Gamma processes, we define a Subsampling Markov Random Field (MRF) \cite{kindermann1980markov,li1995markov} to constrain the $q_k^d$ of all documents,

\begin{definition}[Subsampling Markov Random Field] The subsampling probabilities of all the documents on a component/topic in the global Gamma process have the following constraint,
\begin{equation}
\begin{aligned}
p(\{q_k^d\}_{d=1}^D) &= \prod_{C \in clique(Network)} \psi(C)
\\
&= \frac{1}{Z(q)} \exp \left( -\sum_{<d_i,d_j> \in C} \|q_{d_i} - q_{d_j}\|^2  \right )
\end{aligned}
\end{equation}
where $Network$ is the document network, $\psi(C)$ is the energy function of MRF and $Z(q)$ is the normalization part and also called partition function.
\end{definition}

Through this subsampling MRF constraint, the marginal distribution of each subsampling probability $q_k^d$ dependents on the values of its neighbors. Therefore, the $q_k^d$ of linked documents will be similar, which ensures the proposed NRT achieve Property \ref{p1}.
\begin{figure}[t]
\centering
    \tikzstyle{rv}=[circle,
                        thick,
                        minimum size=0.8cm,
                        draw=black!80,
                        ]
    \tikzstyle{line}=[->,
                        solid,
                        line width=1pt,
                        draw=black!80,
                        ]
    \tikzstyle{dt}=[circle,
                        thick,
                        minimum size=0.8cm,
                        draw=black!80,
                        fill=gray
                        ]
    \begin{tikzpicture}[font=\normalsize,scale=0.5]

    	\node[rv] (1) at (0.3, 0.8) {$a_0, c_0$};

    	\draw[thick, rounded corners] (-7,-9.3) rectangle (7.7,-1);

        \node at (-7,-8.5) [right] {$K$};
    	
        \draw[fill=gray,opacity=.3,very thin,line join=round]
             (-4,-1.5) --
             (7.4,-1.5) --
             (5,-6.5) --
             (-6.5,-6.5) --cycle ;

        \node at (-4.3,-1.8) [right] {\small \emph{Network}};

        \node[rv] (21) at (-4.5, -5.2)  {$q_{1,k}$};
        \node[rv] (22) at (-1, -3)      {$q_{2,k}$};
        \node[rv] (23) at (0.9, -4)     {$q_{3,k}$};
        \node[rv] (24) at (3.3, -5.5)   {$q_{4,k}$};
        \node[rv] (25) at (5.5, -2.8)   {$q_{d,k}$};

        \node[rv] (31) at (-4.5, -8) {$r_{1,k}$};
        \node[rv] (32) at (-1, -8) {$r_{2,k}$};
        \node[rv] (33) at (0.9, -8) {$r_{3,k}$};
        \node[rv] (34) at (3.3, -8) {$r_{4,k}$};
        \node[rv] (35) at (5.5, -8) {$r_{d,k}$};

        \node[rv] (41) at (-4.5, -15.4) {$\Gamma_{1}$};
        \node[rv] (42) at (-1, -15.4) {$\Gamma_{2}$};
        \node[rv] (43) at (0.9, -15.4) {$\Gamma_{3}$};
        \node[rv] (44) at (3.3, -15.4) {$\Gamma_{4}$};
        \node[rv] (45) at (5.5, -15.4) {$\Gamma_{d}$};

        \node[rv] (8) at (-6, -11) {$\Gamma_{0}$};
        \node[rv] (9) at (-8, -11) {$\alpha$};
        \node[rv] (10) at (-7, -13.5) {$H$};

        \draw[thick, rounded corners] (-7,-16.5) rectangle (7.5,-23.5);

        \node at (-7,-22.5) [right] {$K$};

        \draw[thick, rounded corners] (-5.6,-17) rectangle (-3.4,-20.5);
        \draw[thick, rounded corners] (-2.1,-17) rectangle (-0.1,-20.5);
        \draw[thick, rounded corners] (0,-17) rectangle (1.9,-20.5);
        \draw[thick, rounded corners] (2.2,-17) rectangle (4.35,-20.5);
        \draw[thick, rounded corners] (4.36,-17) rectangle (6.6,-20.5);

        \node at (-5.6,-20) [right] {\small $N_1$};
        \node at (-2.3,-20) [right] {\small $N_2$};
        \node at (0.7,-20) [right] {\small $N_3$};
        \node at (2.1,-20) [right] {\small $N_4$};
        \node at (5.4,-20) [right] {\small $N_d$};

        \node[dt] (51) at (-4.5, -18.5) {\small $w^n_{1,k}$};
        \node[dt] (52) at (-1, -18.5) {\small $w^n_{2,k}$};
        \node[dt] (53) at (0.9, -18.5) {\small $w^n_{3,k}$};
        \node[dt] (54) at (3.3, -18.5) {\small $w^n_{4,k}$};
        \node[dt] (55) at (5.5, -18.5) {\small $w^n_{d,k}$};

        \node[rv] (61) at (-4.5, -22) {\small $\beta_{1,k}$};
        \node[rv] (62) at (-1, -22) {\small $\beta_{2,k}$};
        \node[rv] (63) at (0.9, -22) {\small $\beta_{3,k}$};
        \node[rv] (64) at (3.3, -22) {\small $\beta_{r,k}$};
        \node[rv] (65) at (5.5, -22) {\small $\beta_{d,k}$};

        \node[rv] (7) at (0.5, -25.5) {$b_0$};

         \draw[line width=2pt, draw=black!80]  (21)		    -- (22);
    	\draw[line width=2pt, draw=black!80]	(21)		-- (23);
         \draw[line width=2pt, draw=black!80]   (22)		-- (23);
         \draw[line width=2pt, draw=black!80]   (23)		-- (24);
    	\draw[line width=2pt, draw=black!80]	(25)		-- (23);

    	\path[line]
    		(1)		edge (21)
    		(1)		edge (22)
            (1)		edge (23)
            (1)		edge (24)
    		(1)		edge (25)

            (21)		edge (31)
    		(22)		edge (32)
            (23)		edge (33)
            (24)		edge (34)
    		(25)		edge (35)

            (31)		edge (41)
    		(32)		edge (42)
            (33)		edge (43)
            (34)		edge (44)
    		(35)		edge (45)

            (8)		edge (41)
    		(8)		edge (42)
            (8)		edge (43)
            (8)		edge (44)
    		(8)		edge (45)

            (41)		edge (51)
    		(42)		edge (52)
            (43)		edge (53)
            (44)		edge (54)
    		(45)		edge (55)

            (61)		edge (51)
    		(62)		edge (52)
            (63)		edge (53)
            (64)		edge (54)
    		(65)		edge (55)

            (7)		edge (61)
    		(7)		edge (62)
            (7)		edge (63)
            (7)		edge (64)
    		(7)		edge (65)

            (9)		edge (8)
            (10)    edge (8)

    		;

    \end{tikzpicture}
    \caption{Nonparametric Relational Topic (NRT) Model by dependent thinned Gamma Processes and Markov Random Field (MRF)}
    \label{model1pic}
\end{figure}
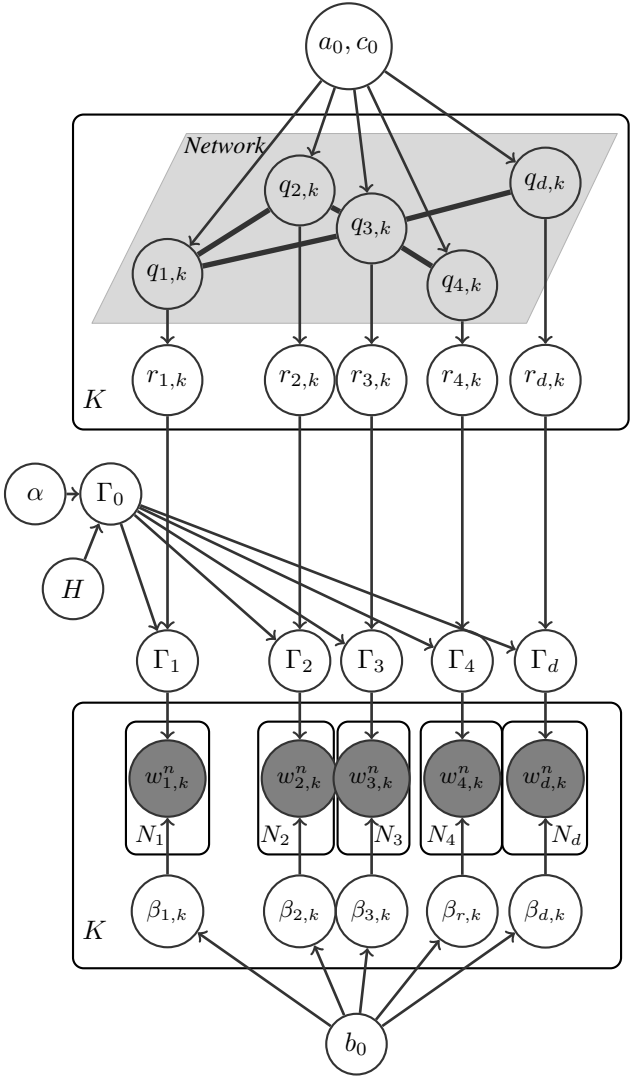

To sum up, the proposed Nonparametric Relational Topic (NRT) Model is,
\begin{equation*}
\begin{aligned}
\Gamma_0 &\sim GaP(\alpha, H)\\
or ~~~\Gamma_0 &= \sum_{k=1}^{\infty} \pi_k \delta_{(x_k, \theta_k)}
\\
p(q_{d,k}) &= beta(q_{d,k}; a_0, c_0) \cdot M(q_{d,k}| q_{-d,-k})
\\
M(q_{d,k}| q_{-d,-k}) &\propto \exp \left( -\sum_{l \in \vartheta(d,k)} \|q_{d,k} - q_{l}\|^2  \right )
\\
r_{d,k} &\sim Ber(q_{d,k})
\\
\Gamma_d &= \sum_{k=1}^{\infty} r_{d,k} \pi_k \delta_{\theta_k}
\end{aligned}
\end{equation*}
With the $\{\Gamma_d\}_{d=1}^D$ for all the documents in hand, the generative procedure of the documents is as follow,
\begin{equation*}
\begin{aligned}
\beta_{d,k} &\sim Gamma(b_0, 1)
\\
w_{d,n,k} &\sim Poisson(\theta_{k,n} r_{d,k} \pi_k \beta_{d,k})
\\
w_{d,n} = \sum_{k=1}^{\infty}w_{d,n,k} &\sim Poisson(\sum_{k=1}^{\infty} \theta_{k,n} r_{d,k} \pi_k \beta_{d,k})
\end{aligned}
\end{equation*}
Considering the relationship between the poisson distribution and the multinomial distribution, the likelihood part is equal to,
\begin{equation*}
\begin{aligned}
z_{d,n, m}  &\sim Discrete(\frac{\theta_{k,n} r_{d,k} \pi_k \beta_{d,k}}{\sum_k \theta_{k,n} r_{d,k} \pi_k \beta_{d,k}})
\\
 m &\in [1, w_{d,n}]
\\
w_{d,n,k} &= \sum_m \delta (z_{d,n, m} = k)
\\
\theta &\sim H
\end{aligned}
\end{equation*}
This form is more convenient for the slice sampling of the model.
Note that the $q$'s are not only with a beta distribution prior but also with a MRF constraint at the same time. We just use this constrain to make the learned $q$'s satisfy the desired property.

\section{Model Inference}

The inference of the proposed (NRT) model is to compute the posterior distribution of latent variables given data,
\begin{equation*}
\begin{aligned}
p(K, \mathbf{\pi}, \mathbf{q}, \mathbf{r}, \mathbf{\theta}, \mathbf{\beta} | \{w_{d,n}\}, Network)
\end{aligned}
\end{equation*}
Here, we use Gibbs sampling method to get samples of this posterior distribution with a truncation (define a relatively big topic number). We also adopt the slice sampling technique \cite{neal2003slice} to develop an exact sampling without the truncation.

\subsection{Gibbs Sampling}

It is difficult to perform posterior inference under infinite mixtures, and a common work-around solution in nonparametric Bayesian learning is to use a truncation method. This method is widely accepted, which uses a relatively big $K$ as the (potential) maximum number of topics.

\textbf{Sampling $q_{d,k}$.}
Since there are additional constraints for the variables $q$, they do not have a closed-formed posterior distribution.

If $r_{d,k}=1$,
\begin{equation}\label{q1}
\begin{aligned}
p(q_{d,k} | \cdots) &\propto q_{d,k}^{a_0 + 1 - 1} (1 - q_{d,k})^{c_0 - 1} \\
&\cdot \exp \left( -\sum_{l \in \vartheta(q_{d,k})} \|q_{d,k} - q_{l}\|^2  \right )
\end{aligned}
\end{equation}

If $r_{d,k}=0$,
\begin{equation}\label{q2}
\begin{aligned}
p(q_{d,k} | \cdots) &\propto q_{d,k}^{a_0 -1 } (1 - q_{d,k})^{c_0 + 1 -1} \\
&\cdot \exp \left( -\sum_{l \in \vartheta(q_{d,k})} \|q_{d,k} - q_{l}\|^2  \right )
\end{aligned}
\end{equation}

Given this conditional distribution of $q_{d,k}$, we can use the efficient \emph{A* sampling} \cite{NIPS20145449} that is developed recently, because the conditional distribution can be decomposed into two parts: $q_{d,k}^{a_0 -1 } (1 - q_{d,k})^{c_0 + 1 -1}$ and $\exp \left( -\sum_{l \in \vartheta(q_{d,k})} \|q_{d,k} - q_{l}\|^2  \right )$. The first part is easily sampled using a beta distribution (proposal distribution), and the second part is a bounded function.

\textbf{Sampling $r_{d,k}$}

\begin{enumerate}
  \item $\forall j, r_{d,j}=0 \rightarrow r_{d,k} = 1$
  \item $\exists n, w_{d,n,k} > 0 \rightarrow r_{d,k} = 1$
  \item $\forall n, w_{d,n,k} = 0$
  \begin{enumerate}
    \item if $\forall n, u_{d,n,k}=0$,
        \begin{equation}
        p(r_{d,k}=1) \propto q_{d,k} \cdot \prod_n pois(0; \theta_{k,n} \pi_k \beta_{d,k})
        \end{equation}
    \item if $\forall n, u_{d,n,k}=0$,
        \begin{equation}
        p^{(1)}(r_{d,k}=0) \propto (1-q_{d,k}) \cdot \prod_n pois(0; \theta_{k,n} \pi_k \beta_{d,k})
        \end{equation}
    \item if $\exists n, u_{d,n,k}>0$,
        \begin{equation}
        \begin{aligned}
        p^{(2)}(r_{d,k}=0) &\propto (1-q_{d,k})\\
        & \cdot \left ( 1 - \prod_n pois(0; \theta_{k,n} \pi_k \beta_{d,k}) \right )
        \end{aligned}
        \end{equation}
  \end{enumerate}
\end{enumerate}

Finally, we can use a discrete distribution to sample $r$ by,
\begin{equation}\label{r}
\begin{aligned}
&~~~~~ p(r_{d,k} = 1 | \cdots) \\
&\propto \frac{p(r_{d,k}=1)}{p(r_{d,k}=1) + p^{(1)}(r_{d,k}=0) + p^{(2)}(r_{d,k}=0)}
\end{aligned}
\end{equation}

\textbf{Sampling $\beta_{d,k}$}
\begin{equation}\label{beta}
\begin{aligned}
p(\beta_{d,k} | \cdots) &\propto Gamma(w_{d\cdot k} + b_0, \frac{1}{r_{d,k} \pi_k + 1})
\end{aligned}
\end{equation}
where $w_{d,\cdot,k} = \sum_n w_{d,n,k}$

\textbf{Sampling $\theta_k$}
\begin{equation}\label{theta}
\begin{aligned}
p(\theta_k | \cdots) &\propto Dir(\alpha_0 + w_{\cdot,1,k},\ldots,\alpha_0 + w_{\cdot,N,k})
\end{aligned}
\end{equation}
where $w_{\cdot,n,k} = \sum_d w_{d,n,k}$

\textbf{Sampling $w_{d,n,k}$ (truncated version)}
\begin{equation}\label{w1}
\begin{aligned}
p(w_{d,n,1},\ldots,w_{d,n,K} | \cdots) &\propto Mult(w_{d,n}; \xi_{d,n,1},\ldots,\xi_{d,n,K})
\end{aligned}
\end{equation}
where
\begin{equation}
\begin{aligned}
\xi_{d,n,k} = \frac{\theta_{k,n} r_{d,k} \pi_k \beta_{d,k}}{\sum_k^K \theta_{k,n} r_{d,k} \pi_k \beta_{d,k}}.
\end{aligned}
\end{equation}

\textbf{Sampling $\pi_{k} $ (truncated version)}
\begin{equation}\label{pi}
\begin{aligned}
p(\pi_{k} | \cdots) &\propto Gamma(1/K + w_{\cdot,\cdot,k}, \frac{1}{\beta_{\cdot,k} + 1})
\end{aligned}
\end{equation}
where $w_{\cdot,\cdot,k} = \sum_d \sum_n w_{d,n,k}$ and $\beta_{\cdot,k} = \sum_d \beta_{d,k}$.

The whole sampling algorithm is summarized in Algorithm \ref{alg1}. Note that the $q_k$ of different $d$ are independent of each other given other variables. So the update of $q_k$ of different $d$ can be implemented in a parallel fashion.

\begin{algorithm}[!t]
\caption{Truncated Version of Gibbs Sampling for NRT}
\label{alg1}
\begin{algorithmic}[1]
    \REQUIRE $Net$, a document network with content $w_{d,n}$
    \ENSURE $K$, $\{\theta_k\}_{k=1}^K$, $\{\pi^d_k\}_{k=1}^K$
    \STATE randomly set initial values for $K$, $\{\theta_k\}_{k=1}^K$, $\{\pi^d_k\}_{k=1}^K$
    \STATE $iter$ = 1;
	\WHILE{ $iter$ $\le$ $max_{iter}$ }
        \FOR{ each topic $k$}
            \FOR{ each document $d$}
                \FOR{ each word $n$ of document $d$}
                    \STATE Update $w_{d,n,k}$ by Eq. (\ref{w1}) ;
                \ENDFOR
                \STATE Update $q_{d,k}$ by Eq. (\ref{q1}) or (\ref{q2}) ;
                \STATE Update $r_{d,k}$ by Eq. (\ref{r}) ;
                \STATE Update $\beta_{d,k}$ by Eq. (\ref{beta}) ;
            \ENDFOR
            \STATE Update $\theta_k$ by Eq. (\ref{theta}) ;
            \STATE Update $\pi_k$ by Eq. (\ref{pi});
        \ENDFOR
    \ENDWHILE
\end{algorithmic}
\end{algorithm}

\subsection{Slice Sampling}

Although the truncated method are commonly accepted in the literature, maintaining a large number of components and their parameters is time and space consuming. An elegant idea (call slice sampling \cite{neal2003slice}) to resolve this problem is to introducing additional variables to adaptively truncate/select the infinite components.

\textbf{Sampling $w_{d,n,k}$ (slice sampling version)}
In order to do slice sampling, sample slice variable as,
\begin{equation}\label{su}
\begin{aligned}
u_{d,n,m} = Unif(0, \zeta_{k})
\end{aligned}
\end{equation}
where fixed positive decreasing sequence $lim_{k \to \infty}\zeta_{k} = 0$. and
\begin{equation}\label{sw}
\begin{aligned}
p(z_{d,n,m} = k | \cdots) &\propto  \xi_{d, n, k} \cdot
\frac{\Pi(u_{d,n,m} \le \zeta_{k})}{\zeta_{k}}
\\
w_{d,n,k} &= \sum_m \delta(z_{d,n,m} = k)
\end{aligned}
\end{equation}
where $lim_{k \to \infty}\zeta_{k} = 0$

\textbf{Sampling $\pi_{k} $ (slice sampling version)}
The construction of Gamma process ($\Gamma_0 \sim GaP(H, \alpha)$) is,
\begin{equation}\label{1}
\begin{aligned}
\Gamma_0 = \sum_{k=1}^{\infty} E_k e^{-T_k} \delta_{\theta_k}
\end{aligned}
\end{equation}
where
\begin{equation}\label{1}
\begin{aligned}
&E_k \sim Exp(\frac{1}{\alpha}),~~T_k \sim Gamma(d_k, \frac{1}{\alpha}),\\
~~&\sum_{k=1}^{\infty} \mathds{1} (d_k = r) \sim poisson(\gamma),\\
~~&\theta_k \sim H,
~~\gamma = \int_\Omega H
\end{aligned}
\end{equation}

The prior of $\pi_k$ is,
\begin{equation}\label{1}
\begin{aligned}
\pi_k &= E_k e^{-T_k} \sim Exp(\frac{1}{\alpha}) \cdot Gamma(d_k, \frac{1}{\alpha})\\
\end{aligned}
\end{equation}
and the posterior is,
\begin{equation}\label{1}
\begin{aligned}
&\pi_k = (E_k, T_k) \\
&\sim poiss(data | E_k e^{-T_k}) \cdot Exp(E_k | \frac{1}{\alpha}) \cdot Gamma(T_k | d_k, \frac{1}{\alpha}) \\
\end{aligned}
\end{equation}

We can sampling this posterior by two gamma distributions,
\begin{equation}\label{slicepi}
\begin{aligned}
p(E_k | T_k) &\sim Gamma(E_k | w_{\cdot,\cdot,k} + 1, \frac{1}{\alpha^{-1} + \beta_{\cdot,k} \cdot e^{-T_k}})
\\
p(T_k | E_k)
&\sim Pois(w_{\cdot,\cdot,k} | \beta_{\cdot,k} \cdot e^{-T_k} E_k )
\\
&~~~~~~~~~~~\cdot Gamma(T_k | d_k, \frac{1}{\alpha}) \\
\end{aligned}
\end{equation}
where $w_{\cdot,\cdot,k} = \sum_d \sum_n w_{d,n,k}$ and $\beta_{\cdot,k} = \sum_d \beta_{d,k}$. The conditional distribution for the indicator $d_k$ is,
\begin{equation}\label{slicedk}
\begin{aligned}
p(d_k = i | \cdots ) \propto p(T_k | d_k = i) \cdot p(d_k = i| \{d_l\}_{l=1}^{k-1})
\end{aligned}
\end{equation}
The second factor is,
\begin{equation}
\begin{aligned}
&p(d_k = i| \{d_l\}_{l=1}^{k-1}) \\
= &\left\{
   \begin{aligned}
   &0           \\
   &~~~~~~~~~~~~~~~~~~~~~~~~~~~~~~~~~~if~~ i < d_{k-1} \\
   &\frac{1 - F(C_{k-1}| \gamma)}
         {1 - F(C_{k-1} - 1| \gamma)} \\
   &~~~~~~~~~~~~~~~~~~~~~~~~~~~~~~~~~~if~~ i = d_{k-1}\\
   &\left (1 -  \frac{1 - F(C_{k-1}| \gamma)}
         {1 - F(C_{k-1} - 1| \gamma)}\right )
         (1 - f(0|\gamma)) f(0|\gamma)^{h-1} \\
   &~~~~~~~~~~~~~~~~~~~~~~~~~~~~~~~~~~if~~ i = d_{k-1} + h\\
   \end{aligned}
   \right.
   \end{aligned}
\end{equation}
where $C_k$ is the number of items in $k$th Poisson process and $C_k \sim Poiss(\gamma)$.

\begin{algorithm}[!t]
\caption{Slice Version of Gibbs Sampling for NRT}
\label{alg2}
\begin{algorithmic}[1]
    \REQUIRE $Net$, a document network with content $w_{d,n}$
    \ENSURE $K$, $\{\theta_k\}_{k=1}^K$, $\{\pi^d_k\}_{k=1}^K$
    \STATE randomly set initial values for $K$, $\{\theta_k\}_{k=1}^K$, $\{\pi^d_k\}_{k=1}^K$
    \STATE $iter$ = 1;
	\WHILE{ $iter$ $\le$ $max_{iter}$ }
        \FOR{ each topic $k$}
            \FOR{ each document $d$}
                \FOR{ each word $n$ of document $d$}
                    \STATE Sample slice variable $u_{d,n,k}$ by Eq. (\ref{su}) ;
                    \STATE Update $w_{d,n,k}$ by Eq. (\ref{sw}) ;
                \ENDFOR
                \STATE Update $q_{d,k}$ by Eq. (\ref{q1}) or (\ref{q2}) ;
                \STATE Update $r_{d,k}$ by Eq. (\ref{r}) ;
                \STATE Update $\beta_{d,k}$ by Eq. (\ref{beta}) ;
            \ENDFOR
            \STATE Update $\theta_k$ by Eq. (\ref{theta}) ;
            \STATE Update $\pi_k$ by Eq. (\ref{slicepi});
            \STATE Update $d_k$ by Eq. (\ref{slicedk});
        \ENDFOR
    \ENDWHILE
\end{algorithmic}
\end{algorithm}

Note that the $d_k$, $E_k$ and $T_k$ are introduced additional variables. They are not in the original model, and their appearances are only for the sampling without the help of the truncation level.
The whole slice sampling algorithm is summarized in Algorithm \ref{alg2}.


\section{Experiments}

In this section, we evaluate the effectiveness of the proposed model in learning the hidden topics from document networks. First, we use a small synthetic dataset to demonstrate the model's ability to recover the number of available topics in the dataset. We then show its usefulness using real-world datasets.

\subsection{Experiments on synthetic data}

We generated synthetic data to explore the NRT's ability to infer the number of hidden topics from the document network.
We chose a set of ground truth numbers symbolised by $K$, $D$ and $W$ that refer to the number of topics, documents and keywords respectively.
Then, we generate the $K$ global topics by the $W$-dimensional Dirichlet distribution parameterized by $\{ \alpha_1, \dots, \alpha_W \}$ where $\alpha_i = 1 \; \forall i$.
Next, we generate the document interests on these topics by the $K$-dimensional Dirichlet distribution parameterized by $\beta_1, \dots, \beta_K \} \; \forall \beta_i = 1$. Now that we have the topics and the document interests on these topics, we can generate each document as follows: For each document $d$, $N_d$ is chosen to be a number between $\frac{N}{2}$ and $N$.

For each word of document $d$, we firstly draw a topic from the document's interest and then draw a word from the selected topic. Finally, we can obtain a matrix with rows as documents and columns as words, and each entry of this matrix denotes the frequency a particular word in a particular document. The next step is to generate the relations between documents. For each pairs of documents, we compute the inner product between their topic distributions. In order to sparsify these relationships, we only retain the ones where their inner product is greater than 0.2.


\begin{figure*}[!tl]
\centerline{\includegraphics[scale=0.57]{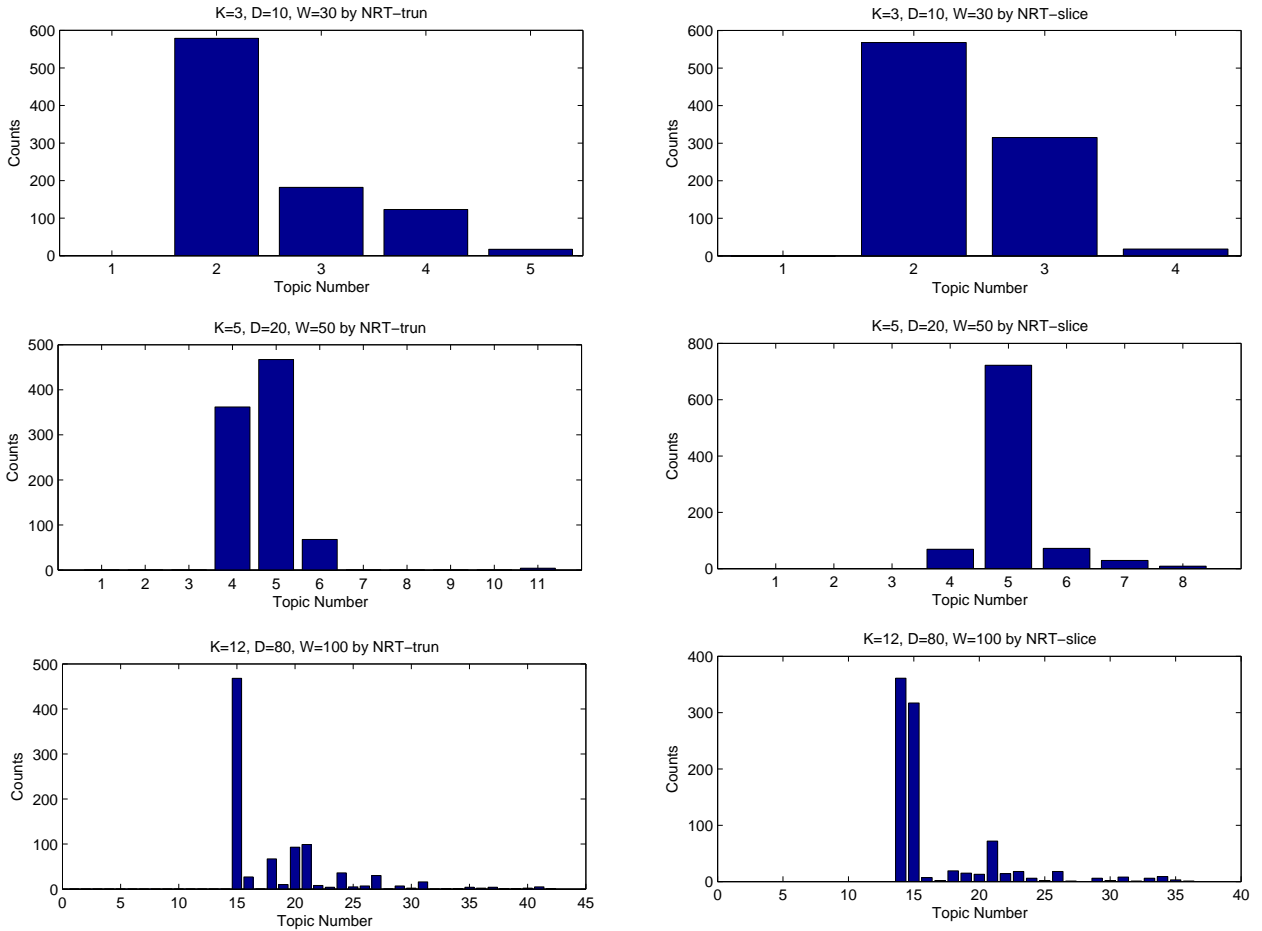}}
\caption{The results of NRT (slice version and truncated version) on synthetic data. The left sub-figures denote the distribution of active topic number from slice version; the right sub-figures denote the distribution of the active topic number from the truncated version. In each sub-figure, the ground-truth of the topic number is given at the top, and the bars represent the frequencies of each possible active topic number.}
\label{fig:syn}
\end{figure*}

Here, we adjust values of $K$, $D$ and $W$ to generate a set of synthetic datasets. The distributions of the learned topic numbers $K$ by the proposed algorithms are shown in Fig. \ref{fig:syn}. The subfiguers in the first column are from the truncated version of the NRT in Algorithm \ref{alg1}, and the subfiguers in the second column are from the slice version of the NRT in Algorithm \ref{alg2}. In each subfigure, the counts of topic numbers from all the iterations (max iteration number is set as 1,000 with 100 burnin) are illustrated in bar charts. Despite the rough initial guess ($K = D \times 10$), we can see that the recovered histogram for $K$ appears to be very similar to the ground truth value with small variance.
The sampled $K$ at plotted across all Gibbs iterations, which shows the Markov Chain begin to mix well after around 400 samples.


\begin{figure}[!t]
\centerline{\includegraphics[scale=0.23]{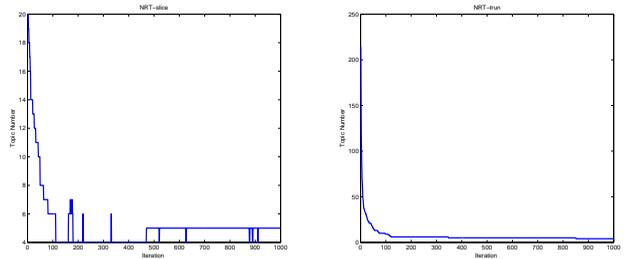}}
\caption{The change of active topic number during the iteration.}
\label{fig:syn2}
\end{figure}


\subsection{Experiments on real-world data}

The real-world datasets used here are:

\begin{itemize}
  \item \textbf{Cora Dataset}\footnote{http://linqs.cs.umd.edu/projects/projects/lbc/} The Cora dataset consists of 2708 scientific publications. The citation network consists of 5429 links. Each publication in the dataset is described by a 0/1-valued word vector indicating the absence/presence of the corresponding word from the dictionary. The dictionary consists of 1433 unique words.
  \item \textbf{Citeseer Dataset} The CiteSeer dataset consists of 3312 scientific publications. The citation network consists of 4732 links. Each publication in the dataset is also described by a 0/1-valued word vector indicating the absence/presence of the corresponding word from the dictionary. The dictionary consists of 3703 unique words \cite{sen:aimag08}.
\end{itemize}

For each dataset, we use 5-fold cross validation to evaluate the performance of the proposed model comparing with Relational Topic Model. The whole dataset is equally split into five parts. At each stage, documents in one part are chosen for testing while the rest of the four parts are used for training. We used the implementation of RTM in Eq. (\ref{rtm1}) from A Fast And Scalable Topic-Modeling Toolbox\footnote{http://www.ics.uci.edu/\~ asuncion/software/fast.htm\#rtm} for comparison.

\begin{table}[t]
\caption{Statistics of Datasets}\label{datasets}
\centering
\begin{tabular}{c|c|c|c}
\hline
~~Datasets~~               & ~\# of documents~   & ~\# of links~   & ~\# of words~\\
\hline
\emph{Cora}                 & 2,708 & 5,429 & 1,433      \\
\hline
\emph{Citeseer}                 & 3,312 & 4,732 & 3,703      \\
\hline
\end{tabular}
\end{table}

In order to quantitatively compare the proposed model with RTM, two evaluation metrics are designed for both real-world datasets: link prediction and document prediction. The link prediction is used to predict the links between test and training documents using learned topics.

The basic idea is that there will be a link between two documents if they have similar interests on topics. The evaluation equation is,
\begin{equation}
\begin{aligned}
Lp = \sum^{Dtest}_{d_i} \sum^{Dtrain}_{d_j} \delta(d_i, d_j)\sum_{n \in d_i} N_n^d * \log( cos(T_{d_j}, W_{n}))
\end{aligned}
\label{eq:lp}
\end{equation}
where $Dtest$ is the number of test documents, $Dtrain$ is the number of training documents, $N_n^d$ is the number of word $n$ in document $d$, and $\delta(d_i, d_j)$ is 1 if there is a link between $d_i$ and $d_j$; 0, otherwise. $T_{d}$ denotes the learned topic distribution of a training document $d$. $W_n$ is the topic distribution (a K-dimensional vector) of a word $n$, which can be evaluated by,
\begin{equation}
\begin{aligned}
W_{n,k} = \frac{\theta_{n,k}}{\sum_l \theta_{n,l}}
\end{aligned}
\end{equation}
where $\{\theta\}_{k=1}^K$ are learned topics. $W_n$ expresses the interest of word $n$ on topics and $T_{d}$ expresses the interest of training document $d$ on topics, so their inner product is used to evaluate the probability of their link. We do not consider the normalization here since it does not influence the comparison made between two models on the same dataset, i.e., a ``max'' operator.

For the word prediction, this basic idea is that a test document has an similar interest on topics with its linked training documents and its words are generated according to its interest. The evaluation equation is,
\begin{equation}
\begin{aligned}
Wp &= \sum^{Dtest}_{d_i} \sum_{n \in d_i} \sum^K_k N_n^d * \log( T_{d_i, k}
\theta_{n,k})
\\
T_{d_i} &= \frac{1}{N_{d_i}}\sum^{Dtrain}_{d_j} \delta(d_i, d_j) T_{d_j}
\end{aligned}
\end{equation}
where $N_{d_i}$ is the number of neighbors that document $i$ has.

\begin{figure*}[!t]
\centerline{\includegraphics[scale=0.57]{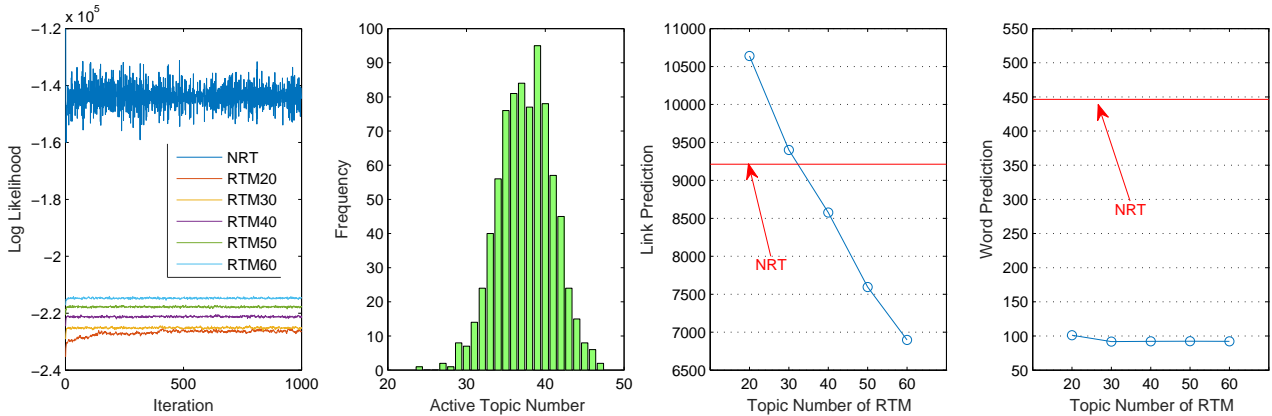}}
\caption{Results of NRT and RTM under different setting ($K=20, 30, 40, 50, 60$) on a first 5-fold of cora dataset. The first subfigure shows the Log-likelihood along iterations; the second subfigure is the learned distribution of active topic number; the third subfigure is the comparison of NRT and RTM on link prediction task; the fourth subfigure is the comparison of NRT and RTM on word prediction task. }
\label{fig:cora1}
\end{figure*}

\begin{figure*}[!t]
\centerline{\includegraphics[scale=0.57]{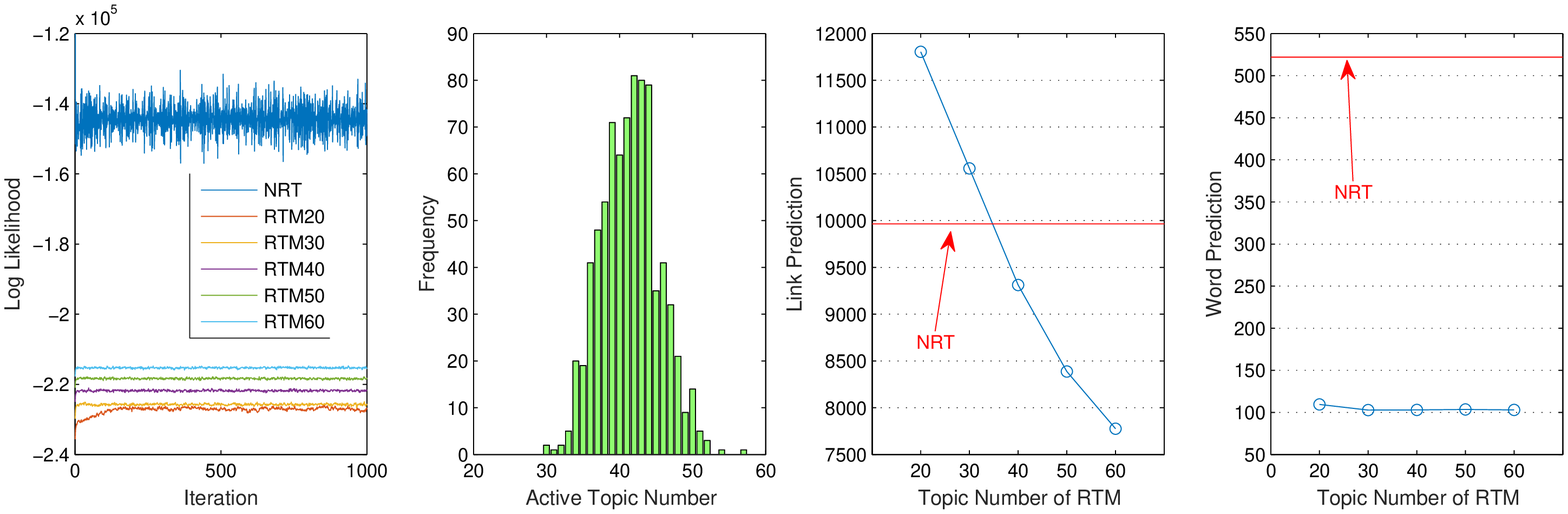}}
\caption{Results of NRT and RTM under different settings ($K=20, 30, 40, 50, 60$) on a second 5-fold of cora dataset.}
\label{fig:cora2}
\end{figure*}

\begin{figure*}[!t]
\centerline{\includegraphics[scale=0.57]{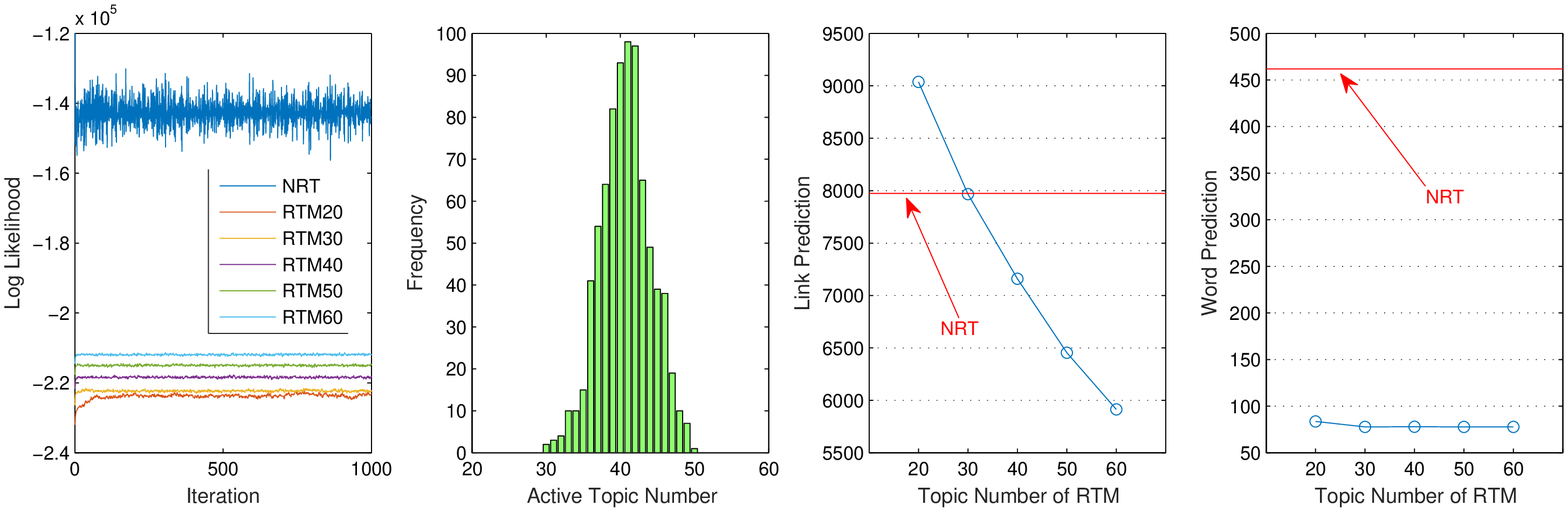}}
\caption{Results of NRT and RTM under different settings ($K=20, 30, 40, 50, 60$) on a third 5-fold of cora dataset.}
\label{fig:cora3}
\end{figure*}

\begin{figure*}[!t]
\centerline{\includegraphics[scale=0.57]{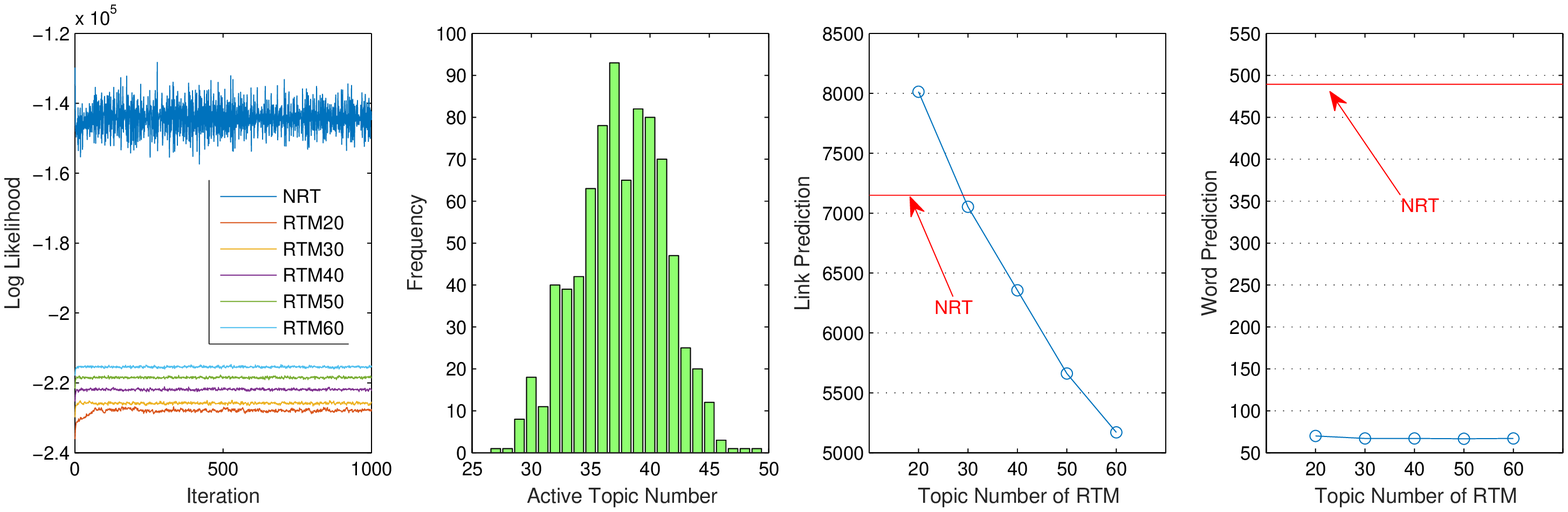}}
\caption{Results of NRT and RTM under different settings ($K=20, 30, 40, 50, 60$) on a fourth 5-fold of cora dataset.}
\label{fig:cora4}
\end{figure*}

\begin{figure*}[!t]
\centerline{\includegraphics[scale=0.57]{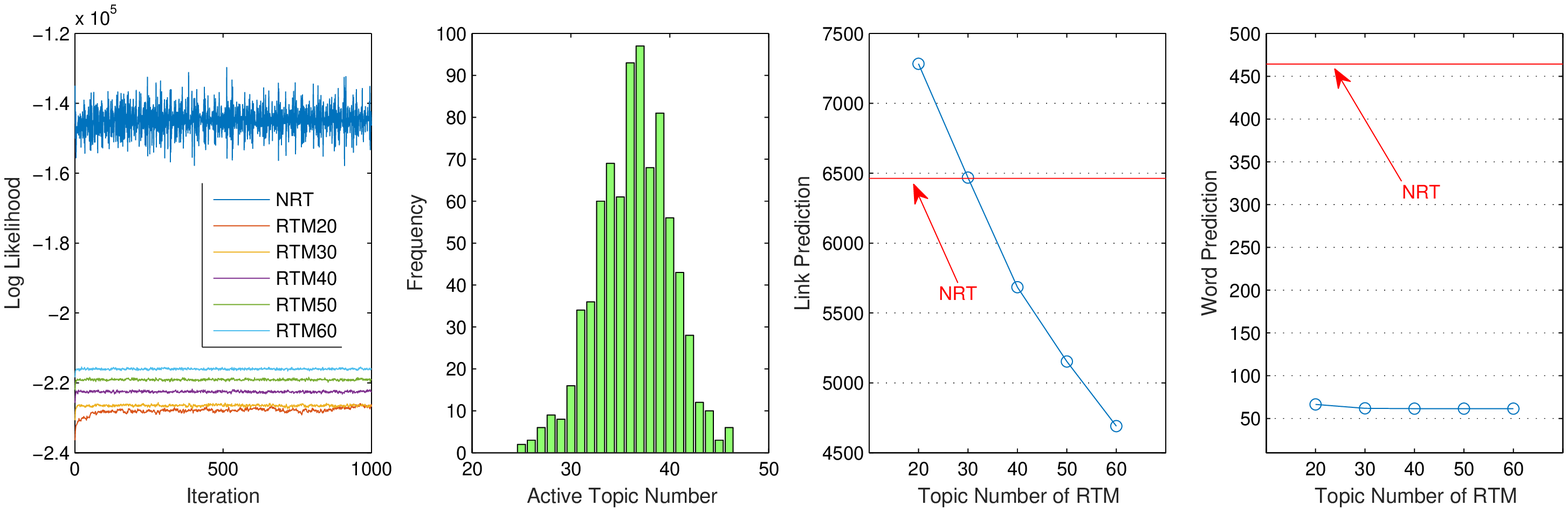}}
\caption{Results of NRT and RTM under different settings ($K=20, 30, 40, 50, 60$) on a fifth 5-fold of cora dataset.}
\label{fig:cora5}
\end{figure*}

\begin{figure*}[!t]
\centerline{\includegraphics[scale=0.57]{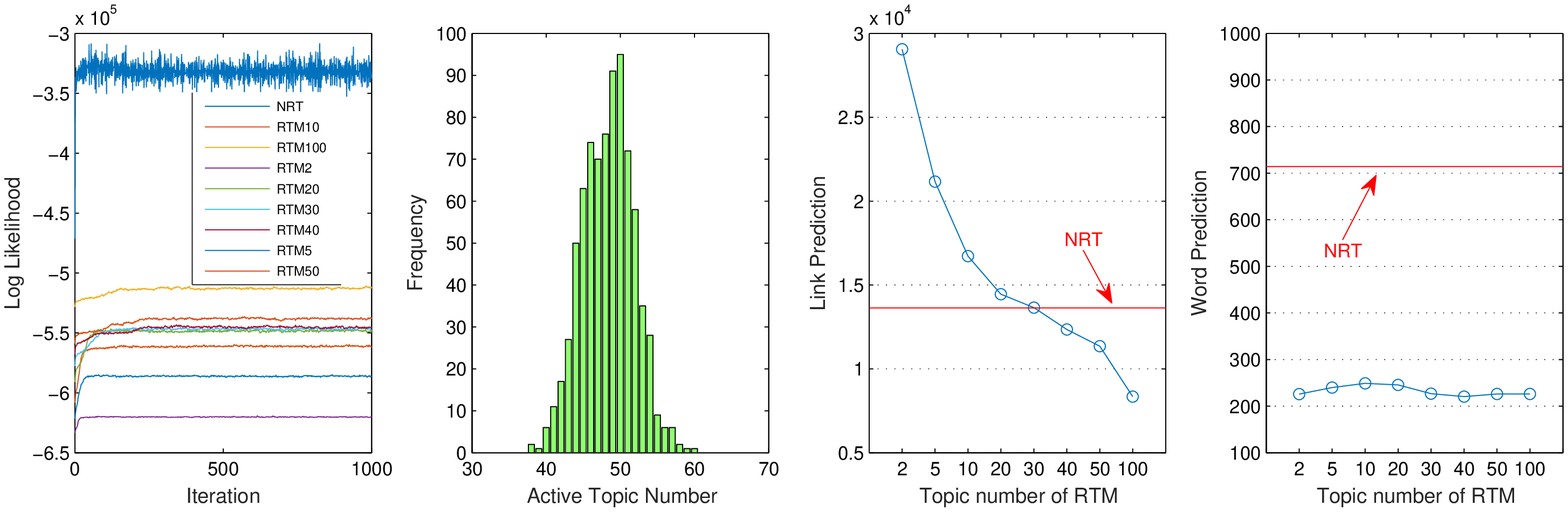}}
\caption{Results of NRT and RTM under different settings ($K=2, 5, 10, 20, 30, 40, 50, 100$) on a first 5-fold of citeseer dataset.}
\label{fig:citeseer1}
\end{figure*}

\begin{figure*}[!t]
\centerline{\includegraphics[scale=0.57]{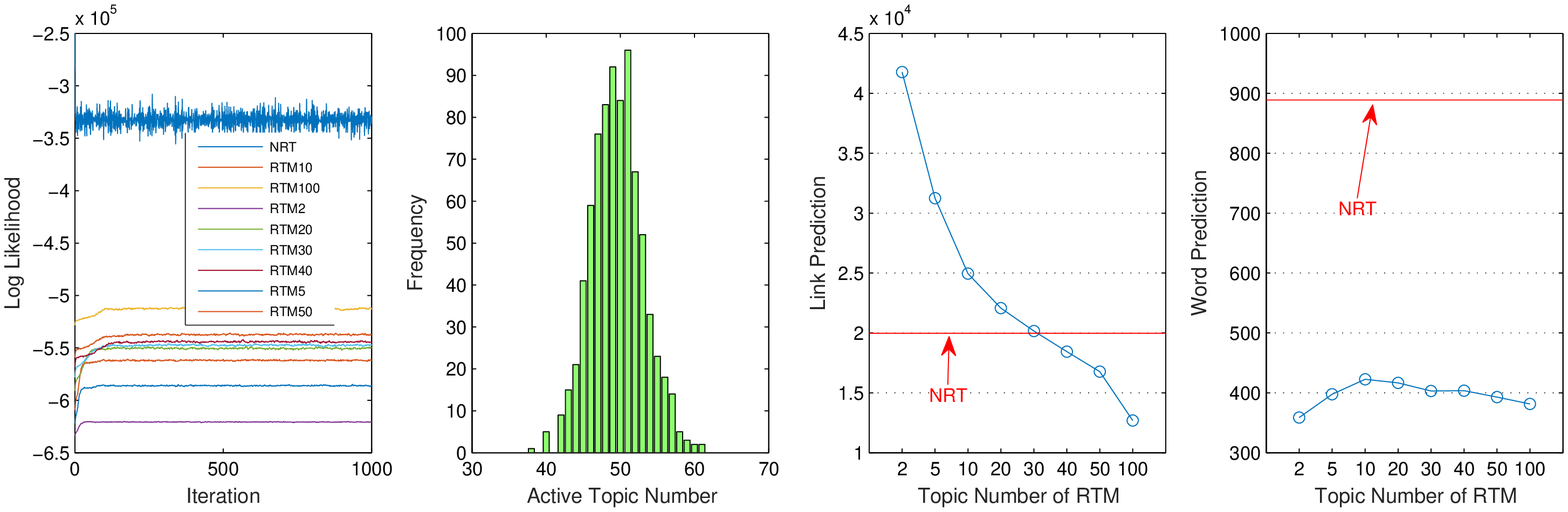}}
\caption{Results of NRT and RTM under different settings ($K=2, 5, 10, 20, 30, 40, 50, 100$) on a second 5-fold of citeseer dataset.}
\label{fig:citeseer2}
\end{figure*}

\begin{figure*}[!t]
\centerline{\includegraphics[scale=0.57]{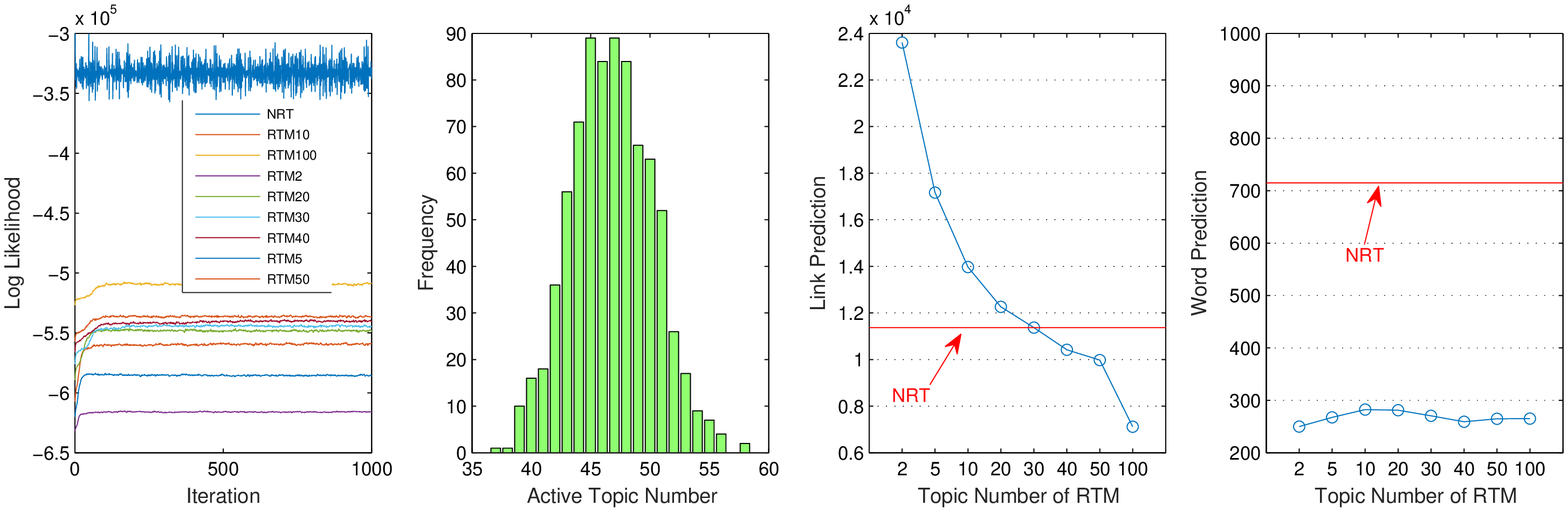}}
\caption{Results of NRT and RTM under different settings ($K=2, 5, 10, 20, 30, 40, 50, 100$) on a third 5-fold of citeseer dataset.}
\label{fig:citeseer3}
\end{figure*}

\begin{figure*}[!t]
\centerline{\includegraphics[scale=0.57]{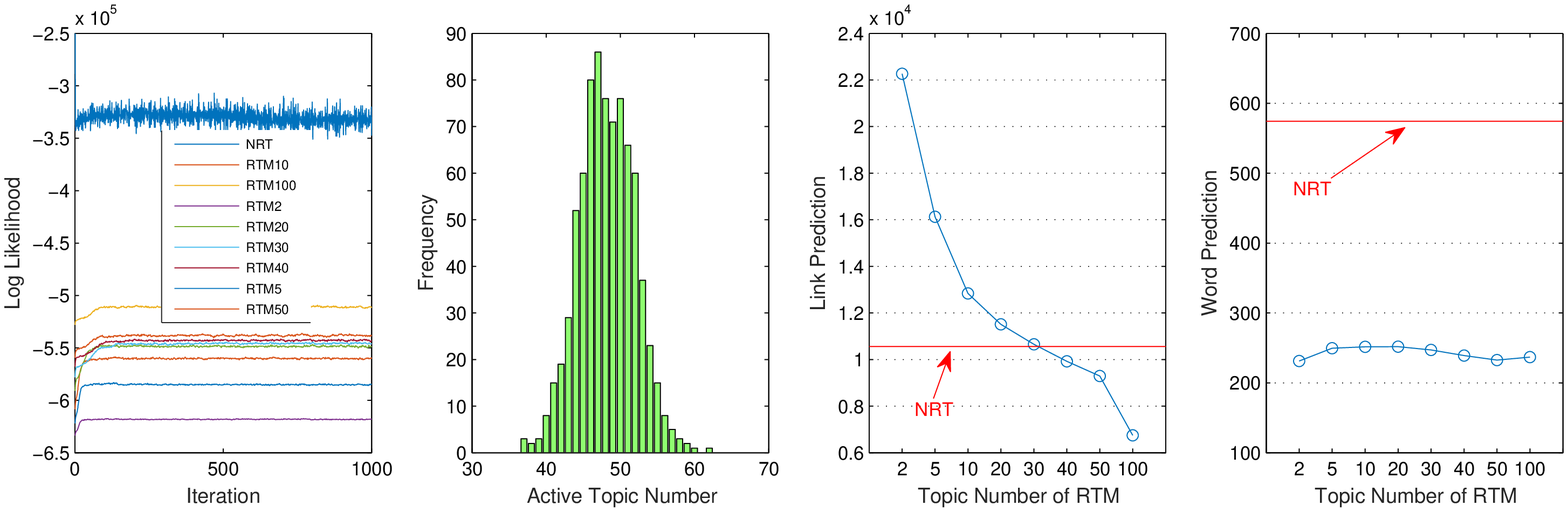}}
\caption{Results of NRT and RTM under different settings ($K=2, 5, 10, 20, 30, 40, 50, 100$) on a fourth 5-fold of citeseer dataset.}
\label{fig:citeseer4}
\end{figure*}

\begin{figure*}[!t]
\centerline{\includegraphics[scale=0.57]{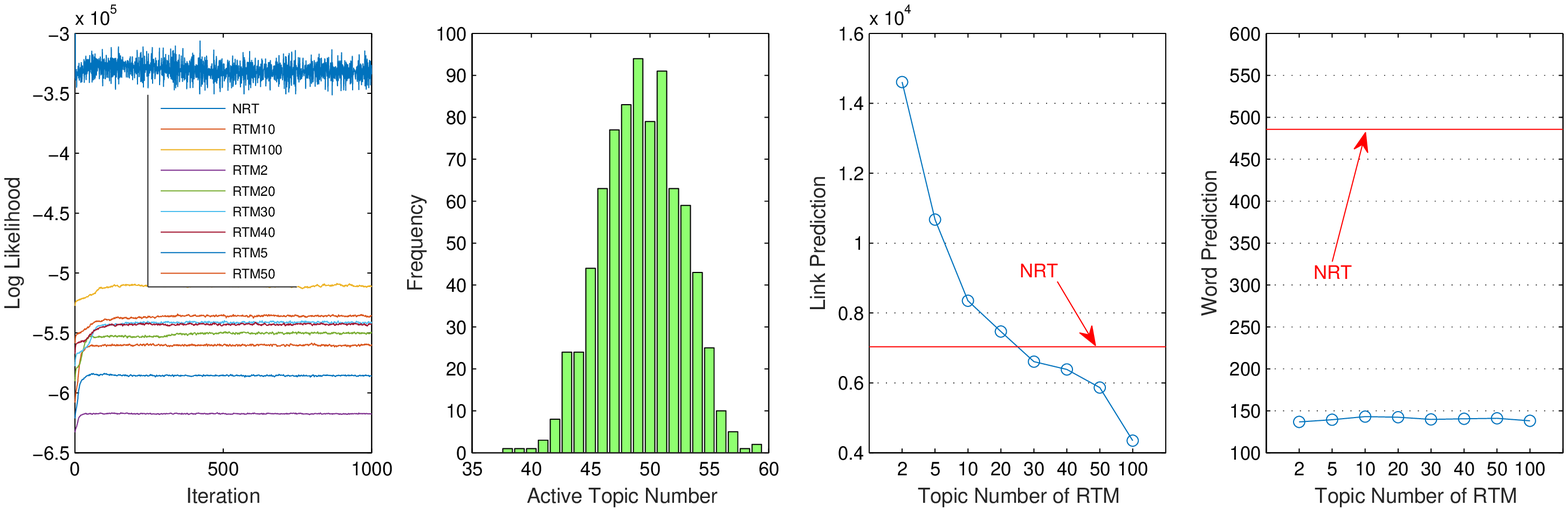}}
\caption{Results of NRT and RTM under different settings ($K=2, 5, 10, 20, 30, 40, 50, 100$) on a fifth 5-fold of citeseer dataset.}
\label{fig:citeseer5}
\end{figure*}

The results on Cora dataset (5-fold) are shown in Fig. \ref{fig:cora1}, \ref{fig:cora2}, \ref{fig:cora3}, \ref{fig:cora4}, and \ref{fig:cora5} and the results on Citeseer dataset (5-fold) are shown in Fig. \ref{fig:citeseer1}, \ref{fig:citeseer2}, \ref{fig:citeseer3}, \ref{fig:citeseer4}, and \ref{fig:citeseer5}, in which we have compared NRT with RTM under several settings. For clarity, we denote RTM with $K = num$ as ``RTM{\it num}''. For example, RTM20 means RTM with $K = 20$.

Note that the slice version of NRT in Algorithm \ref{alg2} is used as the implementation of NRT. The reason is that slice version is more efficient than truncated version because the slice version does not need to keep the (relatively) large number of hidden topics in memory (the initial guess for the number of topics is normally set as larger than the number of documents).

We notice that our algorithm mixed better than some of the $K$ settings in RTM and is generally compatible with the rest.
As shown in left subfigures in each group of data, the likelihood by NRT model is generally larger than the RTM under various settings. It means that the proposed model fits or explains the data better than the RTM.
As with the synthetic case, we also plot the distribution of $K$. We compared our method with RTM in terms of link and word prediction. In terms of word prediction, our algorithm consistently outperform RTM in every category. In terms of link prediction, NRT's performance is not universally better than RTM, where we noticed some less accurate results under some RTM settings. We can see that there is a trend for the link prediction with respective to the topic number in RTM. This trend comes from the evaluation equation \ref{eq:lp}. The RTM with smaller topic number tends to have bigger provability of observed links, which has also been observed in \cite{zhujun2013}. At an extreme situation ($K=1$), the RTM reaches its best performance on link prediction. The problem is how to choose the hidden topic number for RTM. Take Cora dataset as an example. The candidates of possible topic number are at least within $[1, 2708 ]$. However, for the proposed NRT model, the active topic number is automatically learned from the data (for Cora dataset K around $42$). Without any prior domain knowledge, this topic number can achieve relatively good results on the link prediction considering its large range $[1, 2708 ]$.
In terms of overall result, we argue that in the absence of an accurate domain knowledge of $K$ value, the NRT algorithm has allowed us achieving better and more robust performance compared with the current state-of-the-art methods.

\section{Conclusions and future study}

Despite of the success of existing relational topic models in discovering hidden topics from document networks, they are based on the unrealistic assumption, for many real-world applications, that the number of topics can be easily predefined. In order to relax this assumption, we have presented a nonparametric relational topic model. In our proposed model, the stochastic processes are adopted to replace the fixed-dimensional probability distributions used by existing relational topic models which lead to the necessity of pre-defining the number of topics. At the same time, introducing stochastic processes leads to the difficulty with model inference, and we have therefore also presented truncated Gibbs and slice sampling algorithms for the proposed model. Experiments on both the synthetic dataset and the real-world dataset have demonstrated our method's ability to inference the hidden topics and their number.

In the future, we are interested in making the sampling algorithm scalable to large networks by using new network constrain methods instead of MRFs. Current MRF-based methods do not make the inference efficient enough. We believe that the network constraint methods can avoid this issue.

\section*{Acknowledgments}

Research work reported in this paper was partly supported by the Australian Research Council (ARC) under discovery grant DP140101366 and the China Scholarship Council. This work was jointly supported by the National Science Foundation of China under grant no.61471232.

\ifCLASSOPTIONcaptionsoff
  \newpage
\fi

\bibliographystyle{IEEEtran}
\bibliography{TNN}

\begin{thebibliography}{10}
\providecommand{\url}[1]{#1}
\csname url@samestyle\endcsname
\providecommand{\newblock}{\relax}
\providecommand{\bibinfo}[2]{#2}
\providecommand{\BIBentrySTDinterwordspacing}{\spaceskip=0pt\relax}
\providecommand{\BIBentryALTinterwordstretchfactor}{4}
\providecommand{\BIBentryALTinterwordspacing}{\spaceskip=\fontdimen2\font plus
\BIBentryALTinterwordstretchfactor\fontdimen3\font minus
  \fontdimen4\font\relax}
\providecommand{\BIBforeignlanguage}[2]{{%
\expandafter\ifx\csname l@#1\endcsname\relax
\typeout{** WARNING: IEEEtran.bst: No hyphenation pattern has been}%
\typeout{** loaded for the language `#1'. Using the pattern for}%
\typeout{** the default language instead.}%
\else
\language=\csname l@#1\endcsname
\fi
#2}}
\providecommand{\BIBdecl}{\relax}
\BIBdecl

\bibitem{blei2003latent}
D.~M. Blei, A.~Y. Ng, and M.~I. Jordan, ``Latent dirichlet allocation,''
  \emph{Journal of machine Learning Research}, vol.~3, pp. 993--1022, 2003.

\bibitem{Blei:2012:PTM}
\BIBentryALTinterwordspacing
D.~M. Blei, ``Probabilistic topic models,'' \emph{Communications of the ACM},
  vol.~55, no.~4, pp. 77--84, Apr. 2012. [Online]. Available:
  \url{http://doi.acm.org/10.1145/2133806.2133826}
\BIBentrySTDinterwordspacing

\bibitem{6494572}
Z.~Guo, Z.~Zhang, S.~Zhu, Y.~Chi, and Y.~Gong, ``A two-level topic model
  towards knowledge discovery from citation networks,'' \emph{IEEE Transactions
  on Knowledge and Data Engineering}, vol.~26, no.~4, pp. 780--794, April 2014.

\bibitem{klimt2004enron}
B.~Klimt and Y.~Yang, ``The enron corpus: A new dataset for email
  classification research,'' in \emph{Machine learning: ECML 2004}.\hskip 1em
  plus 0.5em minus 0.4em\relax Springer, 2004, pp. 217--226.

\bibitem{park2003hyperlink}
H.~W. Park, ``Hyperlink network analysis: A new method for the study of social
  structure on the web,'' \emph{Connections}, vol.~25, no.~1, pp. 49--61, 2003.

\bibitem{chang2009relational}
J.~Chang and D.~M. Blei, ``Relational topic models for document networks,'' in
  \emph{AISTATS}, 2009, pp. 81--88.

\bibitem{chang2010hierarchical}
J.~Chang, D.~M. Blei \emph{et~al.}, ``Hierarchical relational models for
  document networks,'' \emph{The Annals of Applied Statistics}, vol.~4, no.~1,
  pp. 124--150, 2010.

\bibitem{zhujun2013}
N.~Chen, J.~Zhu, F.~Xia, and B.~Zhang, ``Discriminative relational topic
  models,'' \emph{IEEE Transactions on Pattern Analysis and Machine
  Intelligence}, vol.~PP, no.~99, pp. 1--1, 2014.

\bibitem{teh2006hierarchical}
Y.~W. Teh, M.~I. Jordan, M.~J. Beal, and D.~M. Blei, ``Hierarchical dirichlet
  processes,'' \emph{Journal of the American Statistical Association}, vol.
  101, no. 476, 2006.

\bibitem{mccallum2007topic}
A.~McCallum, X.~Wang, and A.~Corrada-Emmanuel, ``Topic and role discovery in
  social networks with experiments on enron and academic email.'' \emph{Journal
  of Artificial Intelligence Research}, vol.~30, pp. 249--272, 2007.

\bibitem{cha2012social}
Y.~Cha and J.~Cho, ``Social-network analysis using topic models,'' in
  \emph{Proceedings of the 35th International ACM SIGIR Conference on Research
  and Development in Information Retrieval}, ser. SIGIR '12.\hskip 1em plus
  0.5em minus 0.4em\relax New York, NY, USA: ACM, 2012, pp. 565--574.

\bibitem{tuulos2004combining}
V.~Tuulos and H.~Tirri, ``Combining topic models and social networks for chat
  data mining,'' in \emph{Proceedings of IEEE/WIC/ACM International Conference
  on Web Intelligence}, Sept 2004, pp. 206--213.

\bibitem{wang2011dynamic}
E.~Wang, J.~Silva, R.~Willett, and L.~Carin, ``Dynamic relational topic model
  for social network analysis with noisy links,'' in \emph{IEEE Statistical
  Signal Processing Workshop (SSP)}, June 2011, pp. 497--500.

\bibitem{cha2013incorporating}
Y.~Cha, B.~Bi, C.-C. Hsieh, and J.~Cho, ``Incorporating popularity in topic
  models for social network analysis,'' in \emph{Proceedings of the 36th
  International ACM SIGIR Conference on Research and Development in Information
  Retrieval}, ser. SIGIR '13.\hskip 1em plus 0.5em minus 0.4em\relax New York,
  NY, USA: ACM, 2013, pp. 223--232.

\bibitem{mei2008topic}
Q.~Mei, D.~Cai, D.~Zhang, and C.~Zhai, ``Topic modeling with network
  regularization,'' in \emph{Proceedings of the 17th International Conference
  on World Wide Web}, ser. WWW '08.\hskip 1em plus 0.5em minus 0.4em\relax New
  York, NY, USA: ACM, 2008, pp. 101--110.

\bibitem{pathak2008social}
N.~Pathak, C.~DeLong, A.~Banerjee, and K.~Erickson, ``Social topic models for
  community extraction,'' Tech. Rep., 2008.

\bibitem{airoldi2009mixed}
\BIBentryALTinterwordspacing
E.~M. Airoldi, D.~M. Blei, S.~E. Fienberg, and E.~P. Xing, ``Mixed membership
  stochastic blockmodels,'' \emph{Journal of Machine Learning Research},
  vol.~9, pp. 1981--2014, Jun. 2008. [Online]. Available:
  \url{http://dl.acm.org/citation.cfm?id=1390681.1442798}
\BIBentrySTDinterwordspacing

\bibitem{nallapati2008joint}
R.~M. Nallapati, A.~Ahmed, E.~P. Xing, and W.~W. Cohen, ``Joint latent topic
  models for text and citations,'' in \emph{Proceedings of the 14th ACM SIGKDD
  International Conference on Knowledge Discovery and Data Mining}, ser. KDD
  '08.\hskip 1em plus 0.5em minus 0.4em\relax New York, NY, USA: ACM, 2008, pp.
  542--550.

\bibitem{zhu2013scalable}
Y.~Zhu, X.~Yan, L.~Getoor, and C.~Moore, ``Scalable text and link analysis with
  mixed-topic link models,'' in \emph{Proceedings of the 19th ACM SIGKDD
  International Conference on Knowledge Discovery and Data Mining}, ser. KDD
  '13.\hskip 1em plus 0.5em minus 0.4em\relax New York, NY, USA: ACM, 2013, pp.
  473--481.

\bibitem{dietz2007unsupervised}
L.~Dietz, S.~Bickel, and T.~Scheffer, ``Unsupervised prediction of citation
  influences,'' in \emph{Proceedings of the 24th International Conference on
  Machine Learning}, ser. ICML '07.\hskip 1em plus 0.5em minus 0.4em\relax New
  York, NY, USA: ACM, 2007, pp. 233--240.

\bibitem{he2009detecting}
Q.~He, B.~Chen, J.~Pei, B.~Qiu, P.~Mitra, and L.~Giles, ``Detecting topic
  evolution in scientific literature: How can citations help?'' in
  \emph{Proceedings of the 18th ACM Conference on Information and Knowledge
  Management}, ser. CIKM '09.\hskip 1em plus 0.5em minus 0.4em\relax New York,
  NY, USA: ACM, 2009, pp. 957--966.

\bibitem{sun2009itopicmodel}
Y.~Sun, J.~Han, J.~Gao, and Y.~Yu, ``itopicmodel: Information
  network-integrated topic modeling,'' in \emph{The Ninth IEEE International
  Conference on Data Mining}, Dec 2009, pp. 493--502.

\bibitem{liu2009topic}
Y.~Liu, A.~Niculescu-Mizil, and W.~Gryc, ``Topic-link lda: Joint models of
  topic and author community,'' in \emph{Proceedings of the 26th Annual
  International Conference on Machine Learning}, ser. ICML '09.\hskip 1em plus
  0.5em minus 0.4em\relax New York, NY, USA: ACM, 2009, pp. 665--672.

\bibitem{mccullagh1984generalized}
P.~McCullagh, ``Generalized linear models,'' \emph{European Journal of
  Operational Research}, vol.~16, no.~3, pp. 285--292, 1984.

\bibitem{gershman2012tutorial}
S.~J. Gershman and D.~M. Blei, ``A tutorial on bayesian nonparametric models,''
  \emph{Journal of Mathematical Psychology}, vol.~56, no.~1, pp. 1--12, 2012.

\bibitem{blei2007correlated}
D.~M. Blei and J.~D. Lafferty, ``A correlated topic model of science,''
  \emph{The Annals of Applied Statistics}, pp. 17--35, 2007.

\bibitem{seeger2004gaussian}
M.~Seeger, ``Gaussian processes for machine learning,'' \emph{International
  Journal of Neural Systems}, vol.~14, no.~02, pp. 69--106, 2004.

\bibitem{ghosal2010dirichlet}
S.~Ghosal, \emph{The Dirichlet process, related priors and posterior
  asymptotics}.\hskip 1em plus 0.5em minus 0.4em\relax Chapter, 2010, vol.~2.

\bibitem{blackwell1973ferguson}
D.~Blackwell and J.~B. MacQueen, ``Ferguson distributions via polya urn
  schemes,'' \emph{The Annals of Statistics}, pp. 353--355, 1973.

\bibitem{teh2010dirichlet}
Y.~W. Teh, ``Dirichlet process,'' in \emph{Encyclopedia of machine
  learning}.\hskip 1em plus 0.5em minus 0.4em\relax Springer, 2010, pp.
  280--287.

\bibitem{antoniak1974mixtures}
C.~E. Antoniak, ``Mixtures of dirichlet processes with applications to bayesian
  nonparametric problems,'' \emph{The Annals of Statistics}, pp. 1152--1174,
  1974.

\bibitem{rasmussen1999infinite}
\BIBentryALTinterwordspacing
C.~E. Rasmussen, ``The infinite gaussian mixture model,'' in \emph{Advances in
  Neural Information Processing Systems 12}, S.~Solla, T.~Leen, and
  K.~M\"{u}ller, Eds.\hskip 1em plus 0.5em minus 0.4em\relax MIT Press, 2000,
  pp. 554--560. [Online]. Available:
  \url{http://papers.nips.cc/paper/1745-the-infinite-gaussian-mixture-model.pdf}
\BIBentrySTDinterwordspacing

\bibitem{griffiths2004hierarchical}
T.~Griffiths, M.~Jordan, J.~Tenenbaum, and D.~M. Blei, ``Hierarchical topic
  models and the nested chinese restaurant process,'' \emph{Advances in Neural
  Information Processing Systems}, vol.~16, pp. 106--114, 2004.

\bibitem{neal2000markov}
R.~M. Neal, ``Markov chain sampling methods for dirichlet process mixture
  models,'' \emph{Journal of Computational and Graphical Statistics}, vol.~9,
  no.~2, pp. 249--265, 2000.

\bibitem{carin2011variational}
L.~Carin, D.~M. Blei, and J.~W. Paisley, ``Variational inference for
  stick-breaking beta process priors,'' in \emph{Proceedings of the 28th
  International Conference on Machine Learning (ICML-11)}, 2011, pp. 889--896.

\bibitem{gp2014}
A.~Roychowdhury and B.~Kulis, ``Gamma processes,stick-breaking, and variational
  inference,'' \emph{arXiv preprint arXiv:1410.1068}, 2014.

\bibitem{foti2012unifying}
N.~J. Foti, J.~D. Futoma, D.~N. Rockmore, and S.~Williamson, ``A unifying
  representation for a class of dependent random measures,'' in \emph{AISTATS},
  2013, pp. 20--28.

\bibitem{kindermann1980markov}
R.~Kindermann, J.~L. Snell \emph{et~al.}, \emph{Markov random fields and their
  applications}.\hskip 1em plus 0.5em minus 0.4em\relax American Mathematical
  Society Providence, RI, 1980, vol.~1.

\bibitem{li1995markov}
S.~Z. Li, \emph{Markov random field modeling in computer vision}.\hskip 1em
  plus 0.5em minus 0.4em\relax Springer-Verlag New York, Inc., 1995.

\bibitem{neal2003slice}
R.~M. Neal, ``Slice sampling,'' \emph{Annals of Statistics}, pp. 705--741,
  2003.

\bibitem{NIPS20145449}
\BIBentryALTinterwordspacing
C.~J. Maddison, D.~Tarlow, and T.~Minka, ``A* sampling,'' in \emph{Advances in
  Neural Information Processing Systems 27}, Z.~Ghahramani, M.~Welling,
  C.~Cortes, N.~Lawrence, and K.~Weinberger, Eds.\hskip 1em plus 0.5em minus
  0.4em\relax Curran Associates, Inc., 2014, pp. 3086--3094. [Online].
  Available: \url{http://papers.nips.cc/paper/5449-a-sampling.pdf}
\BIBentrySTDinterwordspacing

\bibitem{sen:aimag08}
P.~Sen, G.~M. Namata, M.~Bilgic, L.~Getoor, B.~Gallagher, and T.~Eliassi-Rad,
  ``Collective classification in network data,'' \emph{AI Magazine}, vol.~29,
  no.~3, pp. 93--106, 2008.

\end{thebibliography}

\begin{IEEEbiography}[{\includegraphics[width=1in,height=1.25in,clip,keepaspectratio]{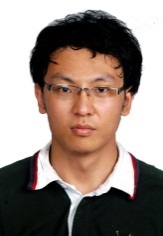}}]{Junyu Xuan}
is received the bachelor's degree in 2008 from China University of Geosciences, Beijing. Currently he is working toward the dual-doctoral degree both in Shanghai University and University of Technology, Sydney. His main research interests include Machine Learning, Complex Network and Web Mining.
\end{IEEEbiography}

\begin{IEEEbiography}[{\includegraphics[width=1in,height=1.25in,clip,keepaspectratio]{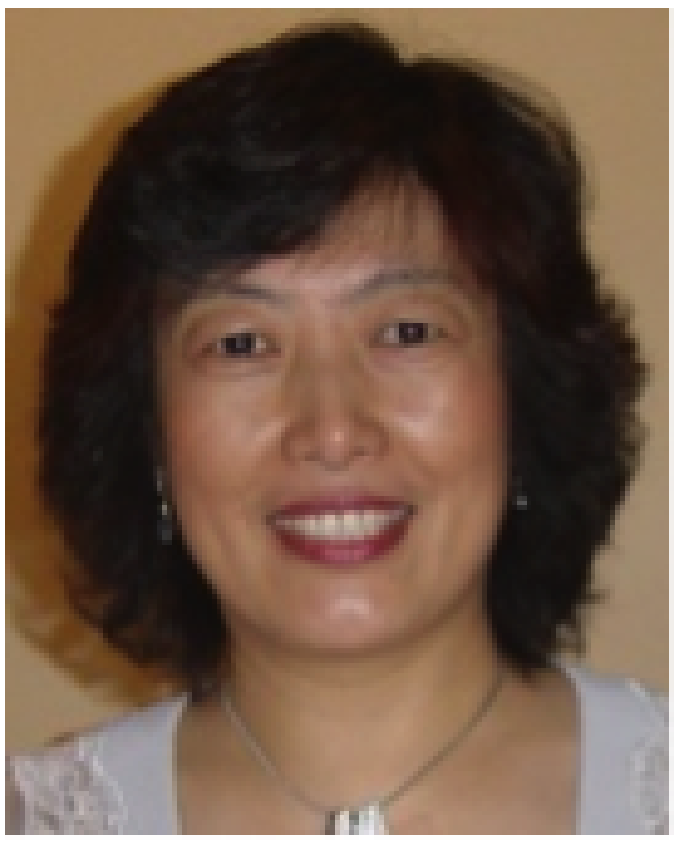}}]{Jie Lu}
is a full professor and Associate Dean in the Faculty of Engineering and Information Technology (FEIT) at the University of Technology, Sydney. Her research interests lie in the area of decision support systems and uncertain information processing. She has published five research books and 270 papers, won five Australian Research Council discovery grants and 10 other grants. She received a University Research Excellent Medal in 2010. She serves as Editor-In-Chief for Knowledge-Based Systems (Elsevier), Editor-In-Chief for International Journal of Computational Intelligence Systems (Atlantis), editor for book series on Intelligent Information Systems (World Scientific) and guest editor of six special issues for international journals, as well as delivered six keynote speeches at international conferences.
\end{IEEEbiography}

\begin{IEEEbiography}[{\includegraphics[width=1in,height=1.25in, clip]{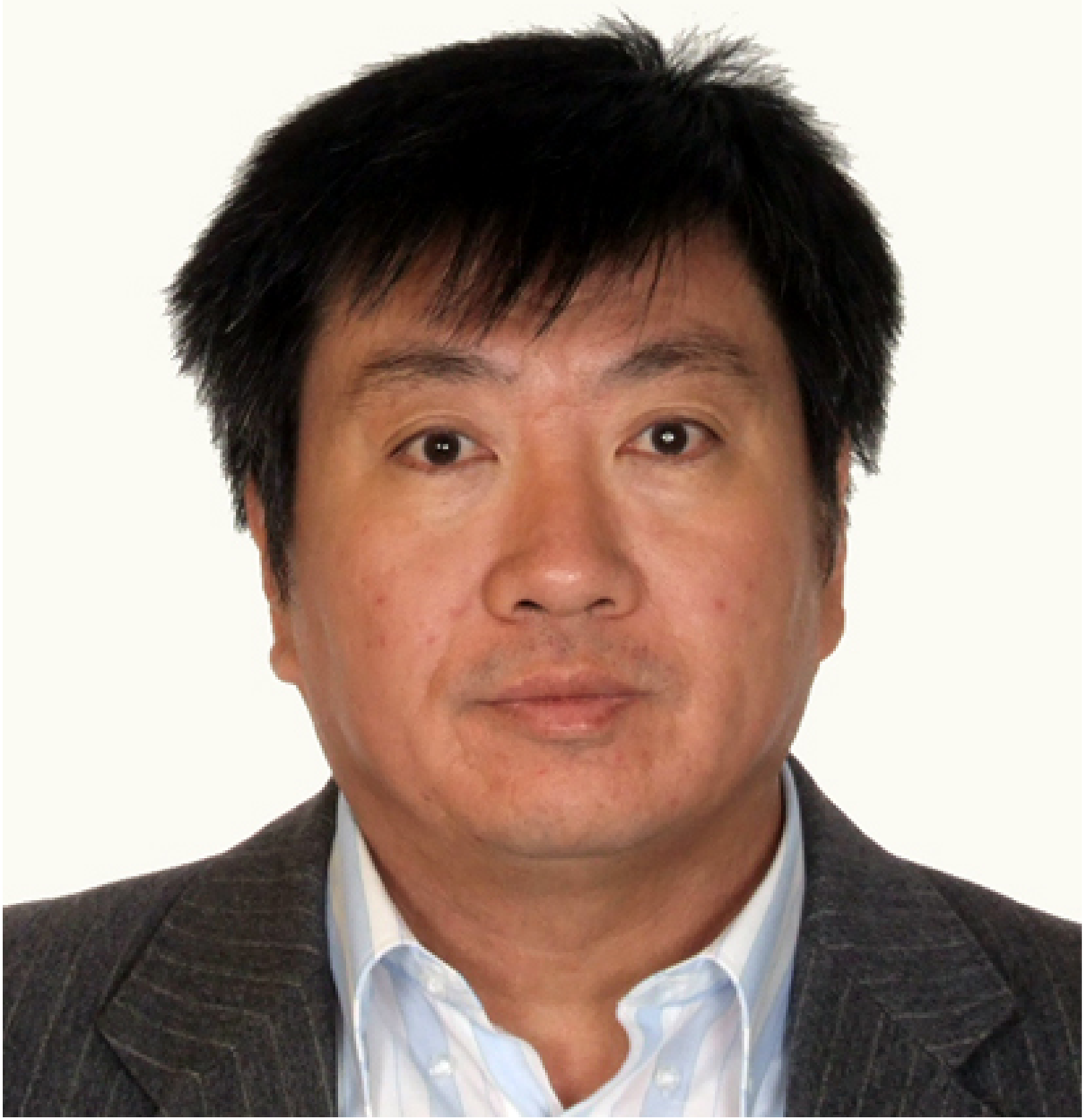}}]{Guangquan Zhang }
is an associate professor in Faculty of Engineering and Information Technology at the University of Technology Sydney (UTS), Australia. He has a PhD in Applied Mathematics from Curtin University of Technology, Australia. He was with the Department of Mathematics, Hebei University, China, from 1979 to 1997, as a Lecturer, Associate Professor and Professor. His main research interests lie in the area of multi-objective, bilevel and group decision making, decision support system tools, fuzzy measure, fuzzy optimization and uncertain information processing. He has published four monographs, four reference books and over 200 papers in refereed journals and conference proceedings and book chapters. He has won four Australian Research Council (ARC) discovery grants and many other research grants.
\end{IEEEbiography}

\begin{IEEEbiography}[{\includegraphics[width=1in,height=1.25in, clip]{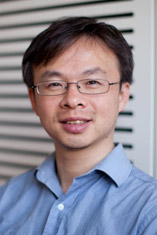}}]{Richard Yi Da Xu}
received the B.Eng. degree in computer engineering from the University of New
South Wales, Sydney, NSW, Australia, in 2001, and the Ph.D. degree in computer sciences from the University of Technology at Sydney (UTS), Sydney, NSW, Australia, in 2006. He is currently a Senior Lecturer with the School of Computing and Communications, UTS. His current research interests include machine learning, computer vision, and statistical data mining.
\end{IEEEbiography}

\begin{IEEEbiography}[{\includegraphics[width=1in,height=1.25in,clip,keepaspectratio]{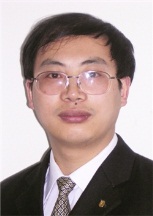}}]{Xiangfeng Luo}
is a professor in the School of Computers, Shanghai University, China. Currently, he is a visiting professor in Purdue University. He received the master's and PhD degrees from the Hefei University of Technology in 2000 and 2003, respectively. He was a postdoctoral researcher with the China Knowledge Grid Research Group, Institute of Computing Technology (ICT), Chinese Academy of Sciences (CAS), from 2003 to 2005. His main research interests include Web Wisdom, Cognitive Informatics, and Text Understanding. He has authored or co-authored more than 50 publications and his publications have appeared in IEEE Trans. on Automation Science and Engineering, IEEE Trans. on Systems, Man, and Cybernetics-Part C, IEEE Trans. on Learning Technology, Concurrency and Computation: Practice and Experience, and New Generation Computing, etc. He has served as the Guest Editor of ACM Transactions on Intelligent Systems and Technology. Dr. Luo has also served on the committees of a number of conferences/workshops, including Program Co-chair of ICWL 2010 (Shanghai), WISM 2012 (Chengdu), CTUW2011 (Sydney) and more than 40 PC members of conferences and workshops.
\end{IEEEbiography}

\end{document}